\documentclass[a4paper,fleqn]{cas-sc}

\usepackage[numbers]{natbib}
\usepackage{algorithmic}
\usepackage[linesnumbered,ruled]{algorithm2e}
\usepackage{threeparttable}
\usepackage{graphicx}
\usepackage{subfigure}
\usepackage{caption}
\usepackage{setspace}
\usepackage[normalem]{ulem}

\usepackage{multirow}

\def\tsc#1{\csdef{#1}{\textsc{\lowercase{#1}}\xspace}}
\tsc{WGM}
\tsc{QE}
\tsc{EP}
\tsc{PMS}
\tsc{BEC}
\tsc{DE}

\begin{document}
\let\WriteBookmarks\relax
\def\floatpagepagefraction{1}
\def\textpagefraction{.001}



\title [mode = title]{Adaptive Convolutional Forecasting Network Based on Time Series Feature-Driven}
\author[inst1]{Dandan Zhang}
\address[inst1]{School of Computer Science and Engineering,
            Southeast University, 
            Nanjing,
            China}
            
\author[inst1]{Zhiqiang Zhang}
\author[inst2]{Nanguang Chen}
\address[inst2]{College of Design and Engineering,
            National University of Singapore, 
            Singapore
            }
\author[inst1]{Yun Wang\corref{cor1}}
\ead{ywang_cse@seu.edu.cn}
\cortext[cor1]{Corresponding author}
\begin{abstract}
Time series data in real-world scenarios contain a substantial amount of nonlinear information, which significantly interferes with the training process of models, leading to decreased prediction performance. Therefore, during the time series forecasting process, extracting the local and global time series patterns and understanding the potential nonlinear features among different time observations are highly significant.
To address this challenge, we introduce multi-resolution convolution and deformable convolution operations. By enlarging the receptive field using convolution kernels with different dilation factors to capture temporal correlation information at different resolutions, and adaptively adjusting the sampling positions through additional offset vectors, we enhance the network's ability to capture potential nonlinear features among time observations. Building upon this, we propose ACNet, an adaptive convolutional network designed to effectively model the local and global temporal dependencies and the nonlinear features between observations in multivariate time series. Specifically, by extracting and fusing time series features at different resolutions, we capture both local contextual information and global patterns in the time series. The designed nonlinear feature adaptive extraction module captures the nonlinear features among different time observations in the time series. We evaluated the performance of ACNet across twelve real-world datasets. The results indicate that ACNet consistently achieves state-of-the-art performance in both short-term and long-term forecasting tasks with favorable runtime efficiency.



\end{abstract}




\maketitle
\doublespacing
\section{Introduction}





Time-series forecasting (TSF) is one of the most critical challenges in time series analysis, with wide-ranging applications in fields such as energy \cite{LIN2024120112,FENG2024120270,BACANIN2023119122}, traffic \cite{QIN2023543}, weather \cite{MO2024120652}, disease \cite{HUANG2024120605}, and more. Its core objective is to leverage past time series data to forecast changing trends over a future time horizon. Effective feature extraction from time series data is essential for improving forecasting accuracy. However, time series data in real-world scenarios often exhibit strong nonlinear features due to factors such as environmental conditions and equipment usage. This poses significant challenges for effective feature extraction. Firstly, there is an increase in noise and uncertainty: data collected by sensors are highly sensitive to environmental changes and external disturbances, leading to increased noise and uncertainty in the time series. Secondly, the extraction of complex dependencies is difficult: time series data in real-world scenarios may exhibit complex nonlinear and dynamic patterns, and the inherent dependency structures and patterns between time observations cannot be accurately captured and described by simple linear models. Additionally, the distribution of time series data in real-world scenarios changes over time, posing higher demands on local feature extraction to effectively capture short-term dynamic behaviors.

\begin{figure}[!htbp]
	\centering
	\subfigure[Raw Time Series (PEMS03).]{
		\includegraphics[width=0.23\columnwidth]{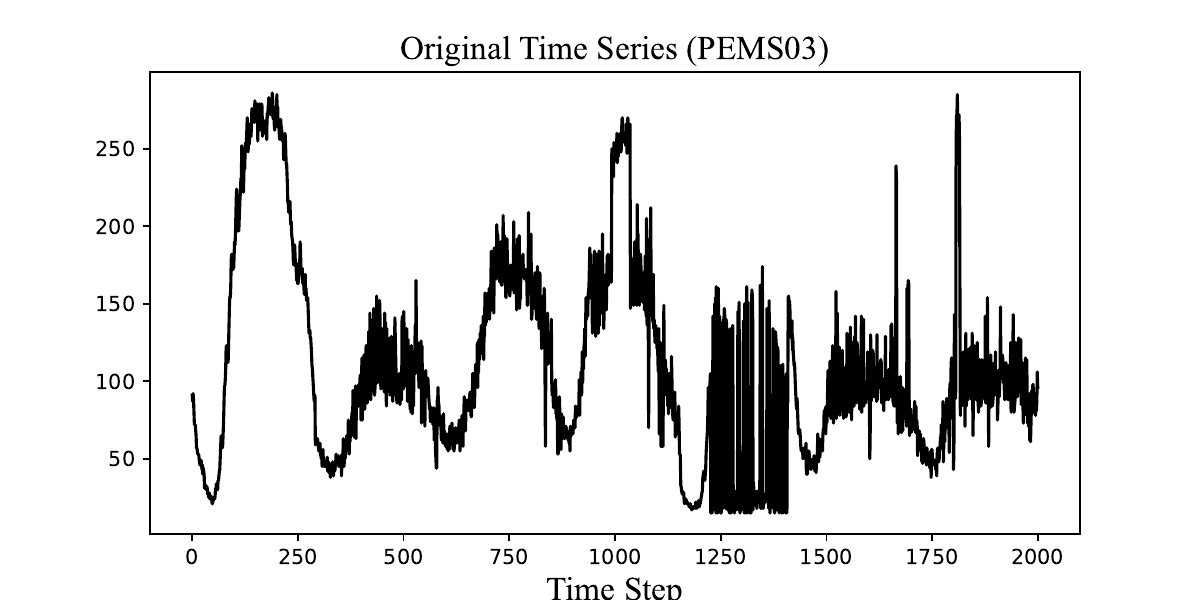}
	}
	\subfigure[Raw Time Series (PEMS04).]{
		\includegraphics[width=0.23\columnwidth]{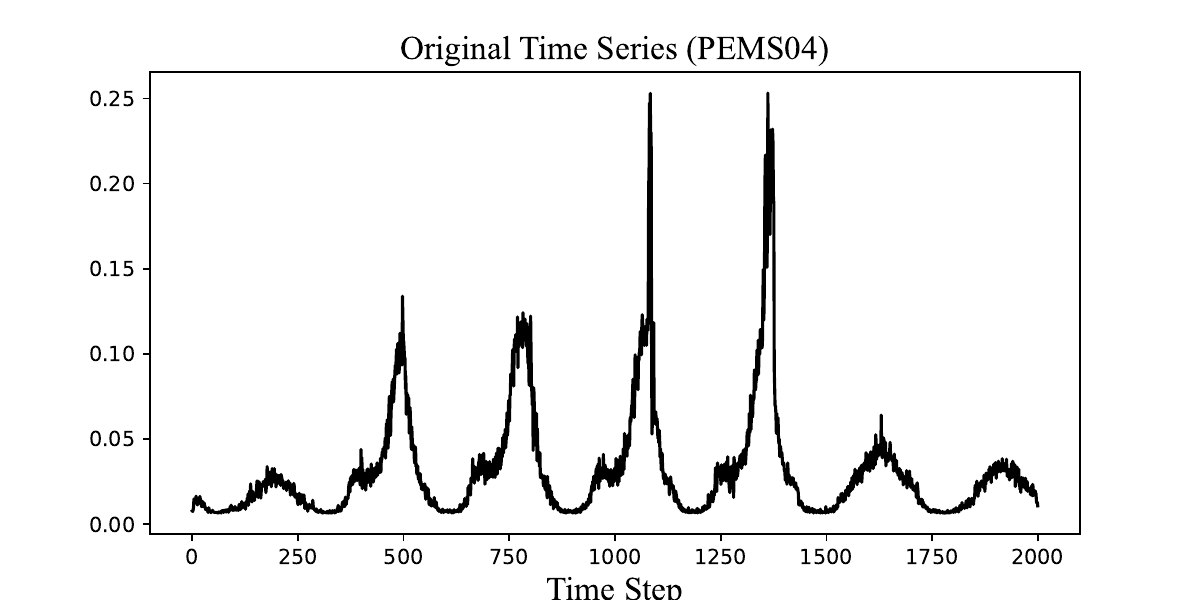}
	}
        \subfigure[Raw Time Series (ETTh2).]{
		\includegraphics[width=0.23\columnwidth]{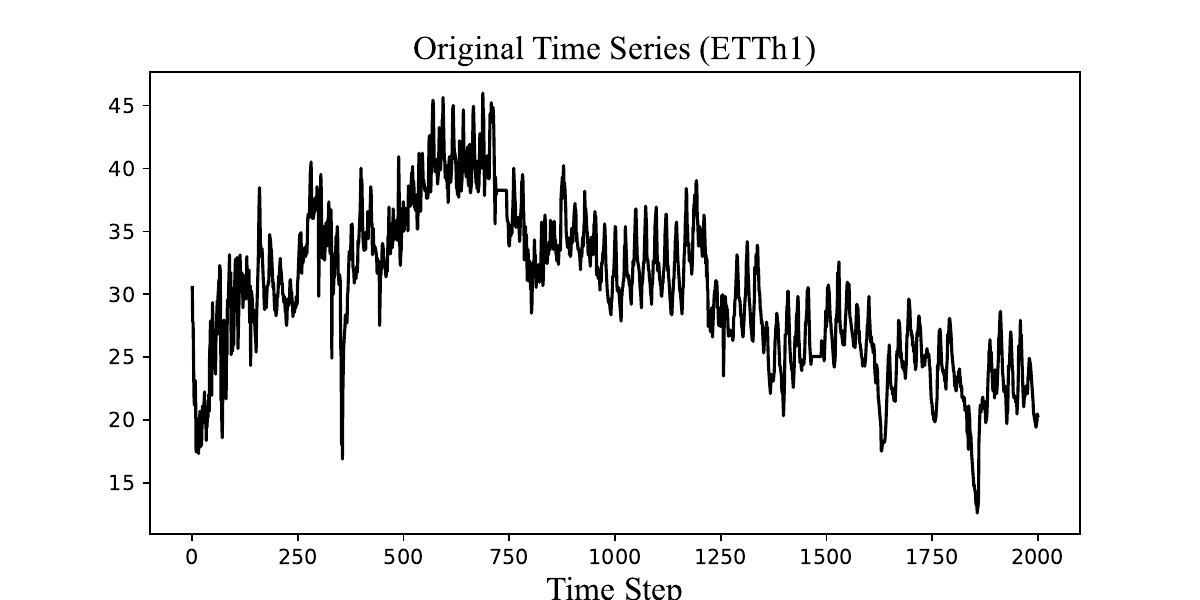}
	}
        \subfigure[Raw Time Series (Exchange).]{
		\includegraphics[width=0.23\columnwidth]{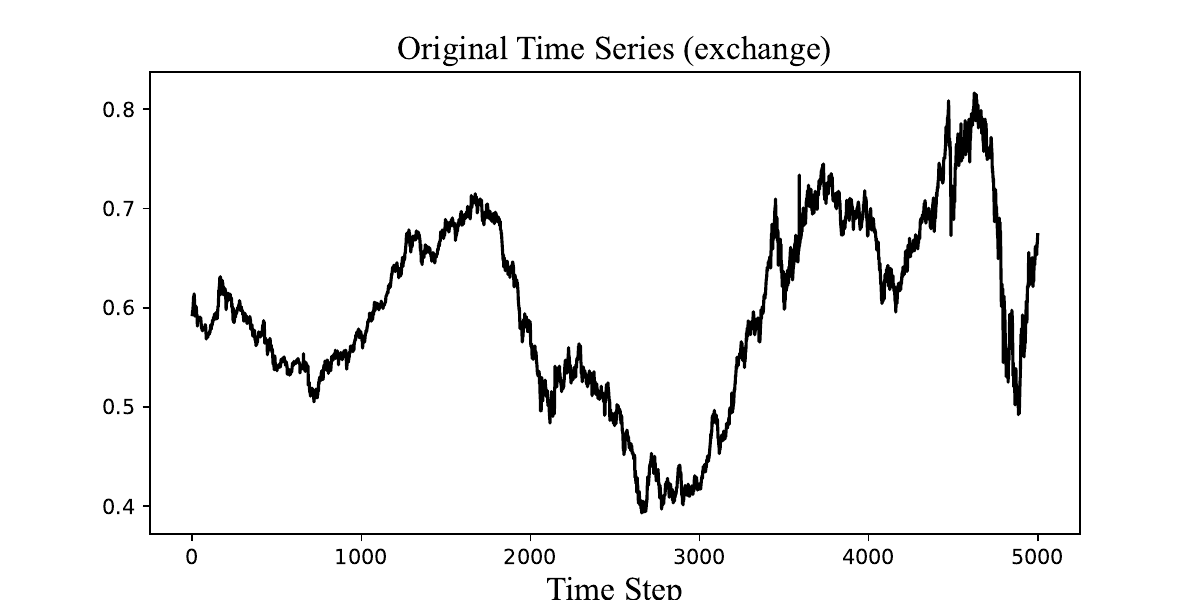}
	}
        \subfigure[Phase Space (PEMS03).]{
		\includegraphics[width=0.23\columnwidth]{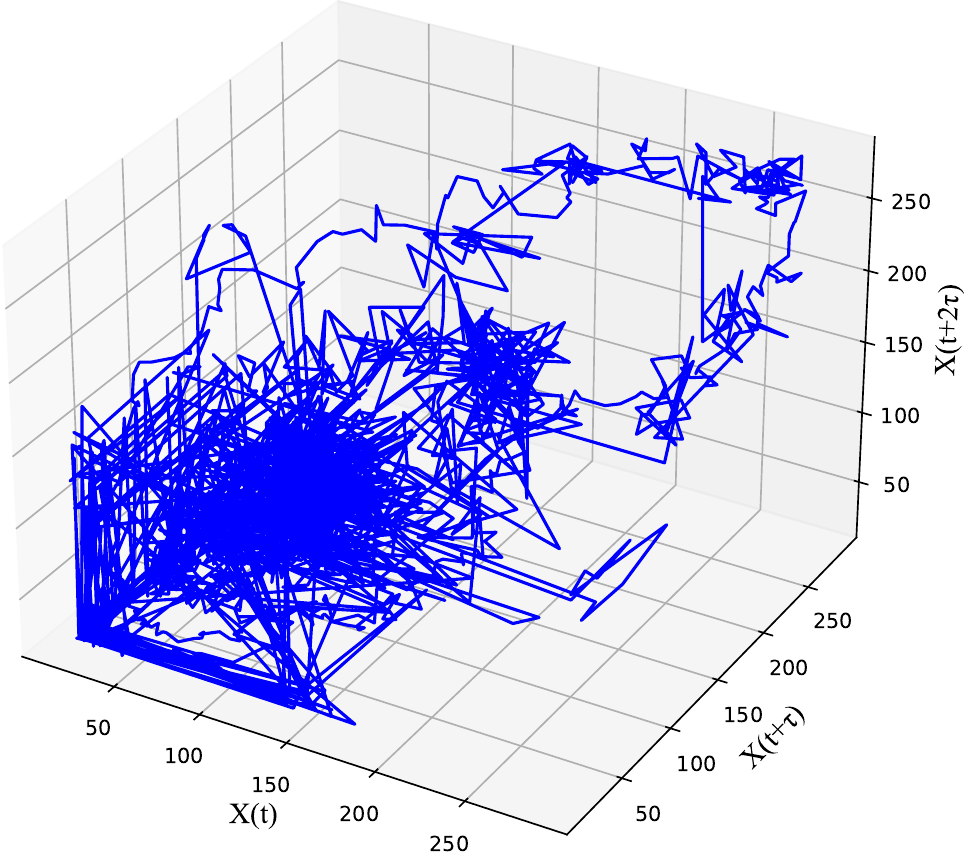}
	}
	\subfigure[Phase Space (PEMS04).]{
		\includegraphics[width=0.23\columnwidth]{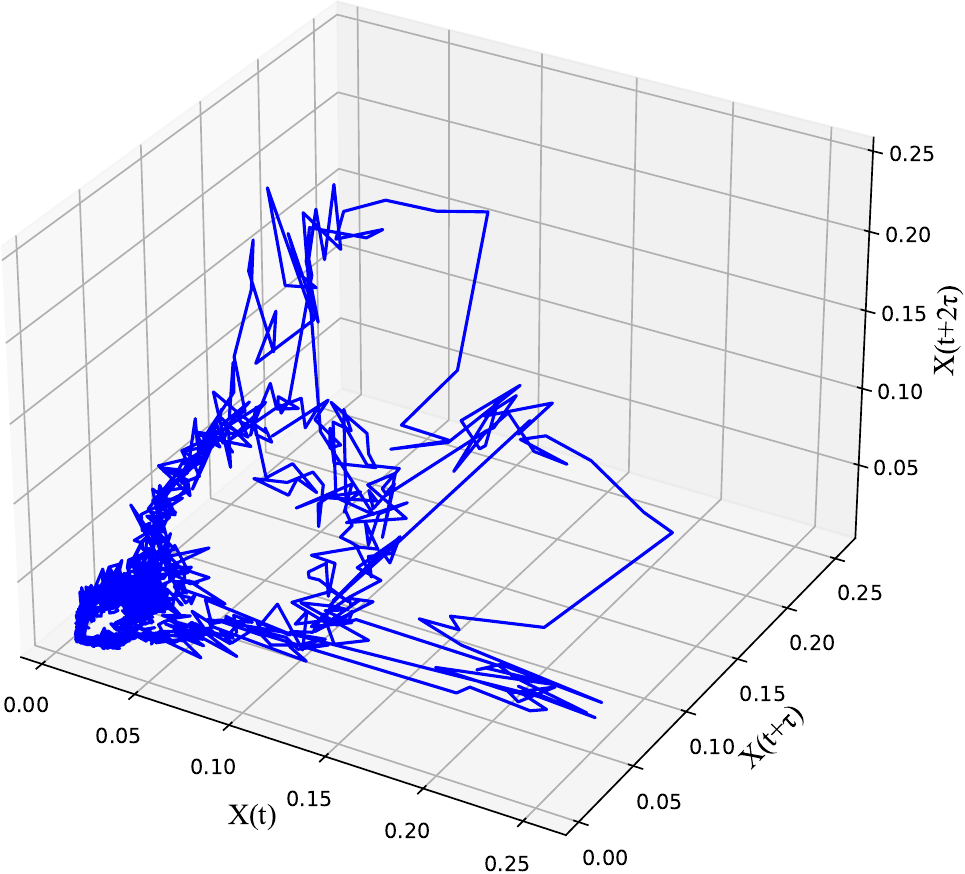}
	}
        \subfigure[Phase Space (ETTh2).]{
		\includegraphics[width=0.23\columnwidth]{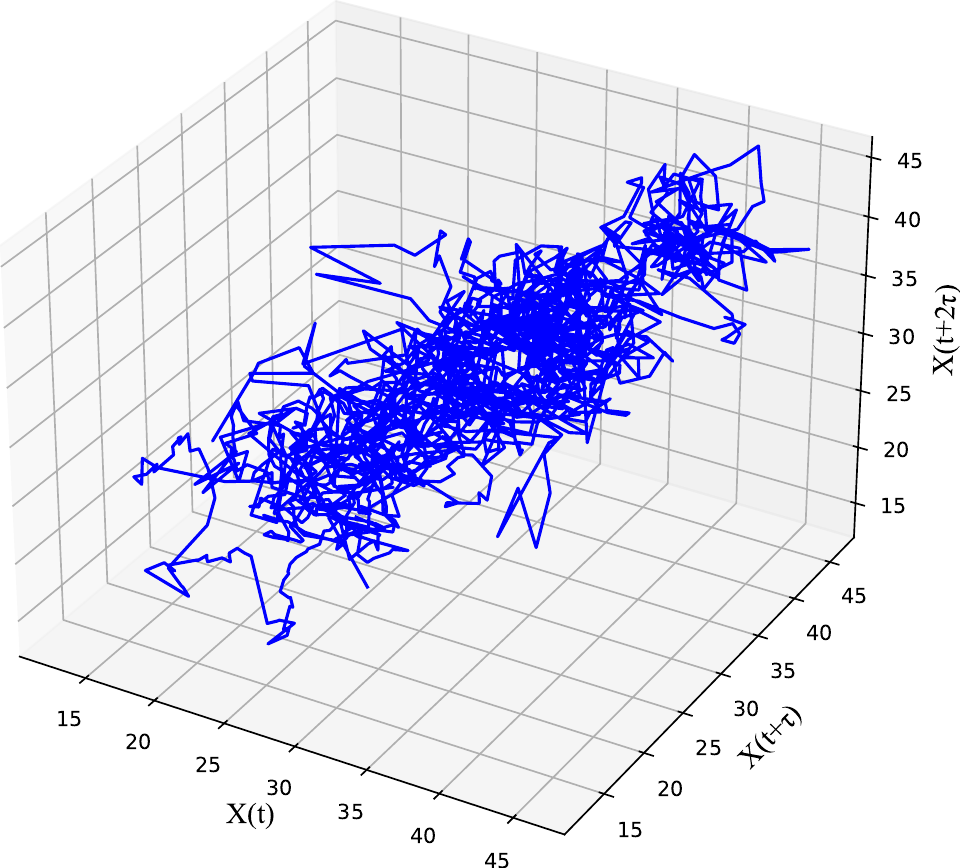}
	}
        \subfigure[Phase Space (Exchange).]{
		\includegraphics[width=0.23\columnwidth]{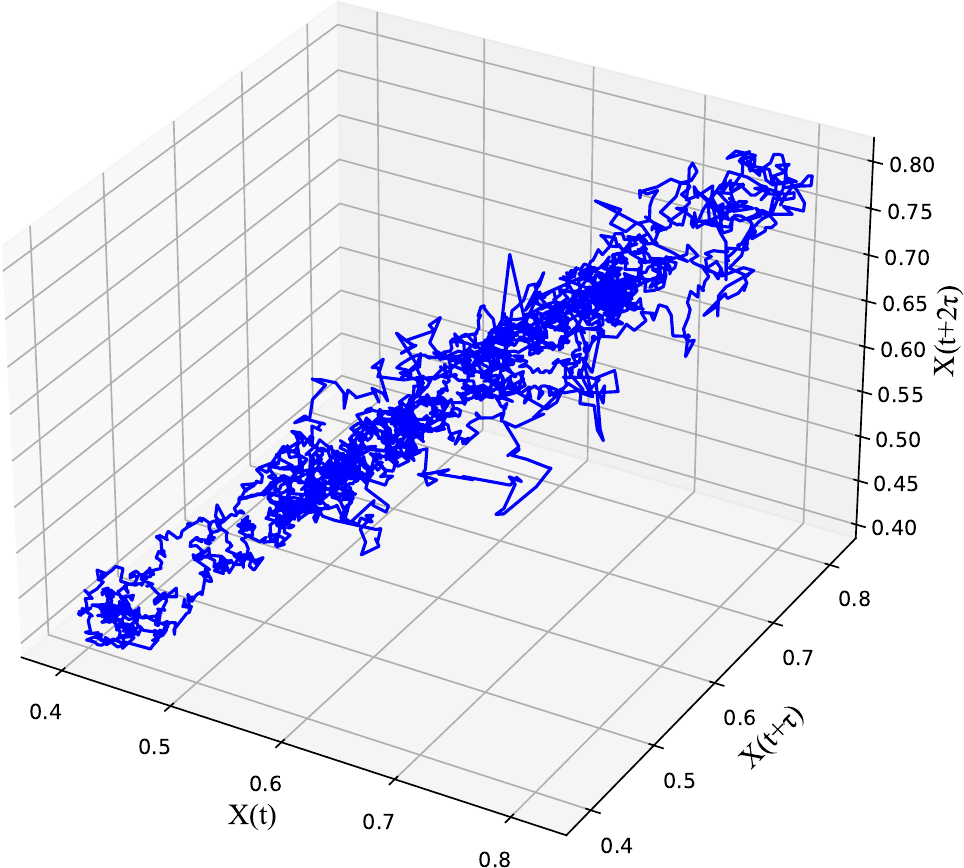}
	}
	\caption{Nonlinear Features of Data.}
	\label{Nonlinear}
\end{figure}

To further illustrate the complex nonlinear characteristics of time series data in real-world scenarios, we performed phase space reconstruction on real datasets from different fields. Figure \ref{Nonlinear} shows the phase space reconstruction diagrams of these datasets, from which the following points can be observed: 1) the trajectory shapes are complex and variable; 2) the trajectories cluster in multiple specific regions; 3) the trajectories are close to each other and intricately intertwined. 
These characteristics indicate that there are strong nonlinear relationships and dynamic behaviors between different time observations in the time series, which greatly limit the effective extraction of time series data features.

To effectively extract complex features and patterns in multivariate time series, research focus on multivariate time series forecasting methods has gradually shifted towards Transformer-based models, which employ multi-encoder-decoder architectures and multi-head self-attention mechanisms. This shift is attributed to the strong capabilities of Transformer models in capturing temporal dependencies in long sequences. Consequently, a series of Transformer-based model-driven approaches have emerged, including PatchTST \cite{nie2023a}, iTransformer \cite{liu2023itransformer}, and Crossformer \cite{hassanin2022crossformer}. 
Thanks to the ability of the self-attention mechanism to extract long-term dependencies in time series, transformer-based models perform well in TSF. However, they often neglect local contextual information. Additionally, DLinear \cite{zeng2023transformers} and the experiments in Section \ref{long} show that transformer-based models are generally ineffective at capturing global temporal correlations in time series. Moreover, using the self-attention mechanism for feature extraction at the core of transformer-based models leads to high computational costs.

The latest trend in TSF research has begun to challenge the efficiency issues of Transformer-based models in prediction tasks. Models like TSMixer \cite{103}, LightTS \cite{20220282303} and DLinear \cite{zeng2023transformers}, which employ simple linear structures, have shown superior performance in TSF compared to the majority of Transformer-based models. However, these models focus on extracting global correlation information in the horizontal time domain, neglecting the understanding of local contextual information and nonlinear information between different time observations, which is crucial for TSF. 

\begin{figure}[h]
\centering
\includegraphics[width=9cm]{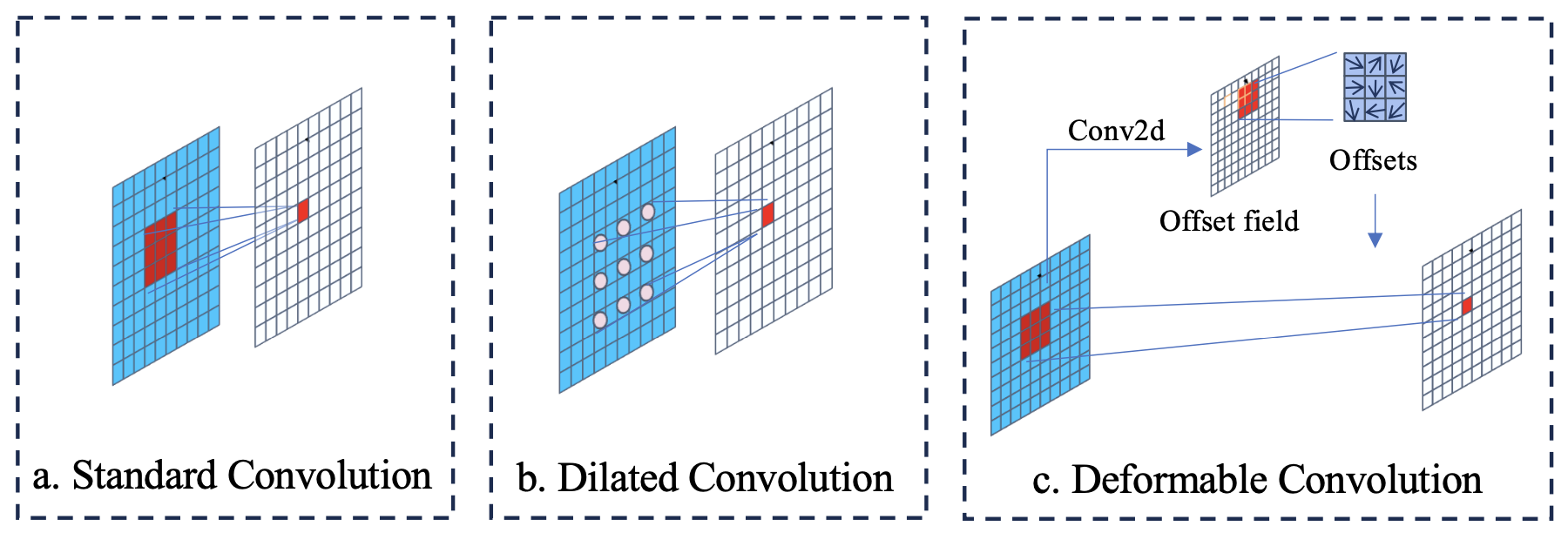}
\caption{Multiple convolution schemes.}
\label{fig:conv}
\end{figure}

Recently, prediction models based on convolutional modules have shown outstanding performance in TSF \cite{cheng2024convtimenet, wang2022micn}. These models typically utilize convolutional layers to extract local nonlinear features from the time series, which to some extent preserve the dynamic information of the time series.
Common methods used for extracting local features include standard convolution and dilated convolution (Figure \ref{fig:conv} a, b). The receptive field of standard convolution typically covers only a fixed-size region within the input sequence, exhibiting regularity and fixedness \cite{WOS:001136775600028}. Dilated convolution \cite{wang2022micn} extends standard convolution by introducing a dilation factor in the convolution kernel, allowing the kernel to span larger receptive fields to capture multi-scale features. However, they possess symmetric receptive fields \cite{WOS:000967248500001}, which limits the ability of convolutional layers to perceive features between asymmetric variables, thus presenting certain limitations in capturing nonlinear characteristics and dynamics of sequences.

To address the aforementioned challenges, we focus on the effective extraction of time series features. Specifically, to capture the temporal domain information of the time series, we consider extracting local features and patterns at different resolutions within the time series. To extract nonlinear features between different time observations in the time series, we improve deformable convolutions \cite{Dai_Qi_Xiong_Li_Zhang_Hu_Wei_2017} (Figure \ref{fig:conv} c), which are commonly used in the field of image processing. Notably, directly applying deformable convolutions to time series prediction tasks may lead to issues of data mismatch and ineffective learning of temporal intervals.

In this work, we propose ACNet, a framework with a multi-resolution mixing architecture capable of extracting local and global information from complex time series data through a temporal feature extraction module (TFE), and then extracting nonlinear features and patterns through a nonlinear feature adaptive extraction module (NFAE). Finally, it predicts future sequences rapidly using a feedforward network and the pseudo-inverse algorithm. With our meticulously designed architecture, ACNet consistently demonstrates leading performance across all experiments, showing superior efficiency in both long-term and short-term forecasting tasks, and covering a wide range of well-established benchmarks. The main contributions are as follows:

\begin{itemize}

\item 
We propose ACNet, an adaptive convolutional network for long-term and short-term time series forecasting. Experimental studies on fifteen real-world datasets demonstrate that the proposed model outperforms the state-of-the-art ConvTimeNet/Crossformer models by 63.4\% and 12.5\% in long-term and short-term time series forecasting, respectively.

\item  
To extract temporal patterns from the time series, we use convolutions with different dilation factors to extract local contextual features at different resolutions of the input sequence. By overlaying feature information of different resolutions at the same position, we ensure the positional invariance of the input sequence. By employing adaptive average pooling operations, the model gains the capability to extract global feature information from the time series, aiding in capturing overall patterns and trends within the sequence.


\item 
To extract nonlinear features and patterns from complex time series data, we designed a nonlinear feature adaptive extraction module. This module employs improved deformable convolutions, which adaptively adjust the shape of the convolutional kernel based on the input time series features, enhancing the model's ability to capture nonlinear characteristics and complex patterns between variables in complex time series, thereby improving the accuracy of time series predictions. Additionally, deformable convolutions allow the network to adaptively modify the convolutional kernel shape according to specific data features, accurately capturing the nonlinear features and patterns between different time observations in the time series, and significantly reducing the extraction of redundant features.

\item 
To improve the computational efficiency of the model, we combine a single hidden layer feedforward neural network (SLFN) with the pseudo-inverse algorithm, which allows for the rapid computation of the optimal solution, significantly reducing training time and computational resource consumption. This approach ensures that our model maintains high prediction accuracy while achieving greater computational efficiency.

\end{itemize}

The remaining structure of this paper is as follows: In Section \ref{sec:sample2}, we introduce existing methods for time series forecasting and the application of deformable convolutions. Section \ref{sec:sample3} provides a detailed description of the ACNet architecture. Section \ref{sec:sample4} presents experimental results validating the effectiveness and efficiency of ACNet. Finally, in Section \ref{sec:sample5}, we provide a comprehensive summary of the predictive research on the ACNet model.

\section{Related work}\label{sec:sample2}
In this section, we delineate the related work into two distinct modules. Section \ref{time-series} provides an overview of prominent contributions within the field of time series forecasting, while Section \ref{deformable} delves into the applications of deformable convolutions in other fields.

\subsection{Time modeling in time series forecasting}\label{time-series}

Deep learning models have achieved significant success in time series forecasting, and they can be primarily categorized into three paradigms: CNN-based models, Transformer-based models, and MLP-based models. 

CNN-based models utilize convolutional kernels along the temporal dimension to effectively capture local temporal patterns.
For instance, TCN \cite{Franceschi_Dieuleveut_Jaggi_2019} uses one-dimensional dilated convolutions to expand the receptive field and performs well in both short-term and long-term forecasting. MICN \cite{wang2022micn} employs multi-scale convolution to extract both global and local contextual relationships. SCINet \cite{20232614296413} decomposes time series into multiple resolutions dynamically through SCI-Blocks in a recursive downsampling fashion to extract features at multiple scales. TSLANet \cite{eldele2024tslanet} learns long-term and short-term relationships in the data through convolutional operations. ConvTimeNet \cite{cheng2024convtimenet} adaptively segments time series into patches and integrates deepwise convolution and pointwise convolution operations to capture global sequence dependencies and cross-variable interactions. However, the aforementioned models are limited to extracting nonlinear features within a fixed receptive field, which hinders their ability to capture nonlinear features among asymmetric variables.


Transformer-based models have garnered widespread attention due to their capability in capturing long-range dependencies through attention mechanisms. For instance, models such as Autoformer \cite{20222412235300}, ETSformer \cite{woo2022etsformer} and FEDformer \cite{zhou2022fedformer} decompose time series data to extract complex patterns. CrossFormer \cite{hassanin2022crossformer} captures temporal and cross-dimensional dependencies in multivariate time series data by first transforming it into a 2D array and then employing two-stage attention layers. PatchTST \cite{nie2023a} employs a channel independence strategy specifically designed to extract correlations of each channel in multivariate data. Pathformer \cite{20240086105} divides the time series into patches of different scales and designs attention mechanisms within and between patches. iTransformer \cite{liu2023itransformer} effectively captures multivariate correlations by independently embedding each variable in the time series as variable sub-tokens. However, as the forecasting horizon increases, the attention mechanism's extraction capability may diminish. 


Models based on MLPs utilize simple linear models to extract abstract representations of time series, such as LightTS \cite{20220282303}, DLinear \cite{zeng2023transformers} decomposes time series and utilizes linear network layers for modeling to achieve prediction. TSMixer \cite{103} combines Patch and MLP, enabling it to extract both temporal correlations and inter-channel correlations. FreTS \cite{Yi2023FrequencydomainMA} and TimeMixer \cite{wang2024timemixer} conduct long-term and short-term forecasting by decoupling historical information of complex patterns in multiscale time series. However, when confronted with high-dimensional data, the expressive capacity of linear networks is constrained, thereby impeding their ability to accurately capture nonlinear features within complex and noisy time series.



\subsection{Application of deformable convolutions}\label{deformable}
Deformable convolutions have found widespread application in the field of image recognition. Recent research has increasingly combined deformable convolutions with attention mechanisms. For instance, in \cite{WOS:000954649600002,WOS:001097273900001}, researchers utilized deformable convolutions to capture significant offset information in specific spatial structures, thereby enhancing recognition precision. In \cite{WOS:001088514600001}, due to the complex geometric transformations and feature blurring present in the data, A2-DCNet modules were employed to capture remote spatial context information from a global perspective. Additionally, \cite{WOS:001116518600002} demonstrated that attention blocks guided by deformable convolutions could acquire semantic information about spatial positions.

Furthermore, research utilizing deformable convolutions in object detection tasks has yielded promising results \cite{WOS:000987415000012}. In \cite{WOS:001142604800050}, a background-guided deformable convolutional autoencoder network was proposed, effectively separating anomalies from complex backgrounds and enhancing anomaly detection capabilities. Moreover, \cite{WOS:001209572500001} showed that stacking deformable convolutions and integrating semantic segmentation improved the understanding of contextual relationships. Finally, \cite{WOS:001164283100017} illustrated that deformable convolutions, akin to spatial attention, could be employed for multiscale feature extraction, while channel attention was used to identify significant features.


To address the aforementioned challenges, we developed a temporal feature extraction module aimed at capturing local contextual information and global patterns within time series data. Additionally, we were inspired by research on the application of deformable convolutions in other domains, prompting us to explore the possibility of introducing them into the field of TSF. By incorporating deformable convolutions, we are able to adaptively capture nonlinear features and patterns within time series data, thereby improving prediction accuracy and performance.

\section{Methodology}\label{sec:sample3}
\begin{figure}[h]
\centering
\includegraphics[width=8.5cm]{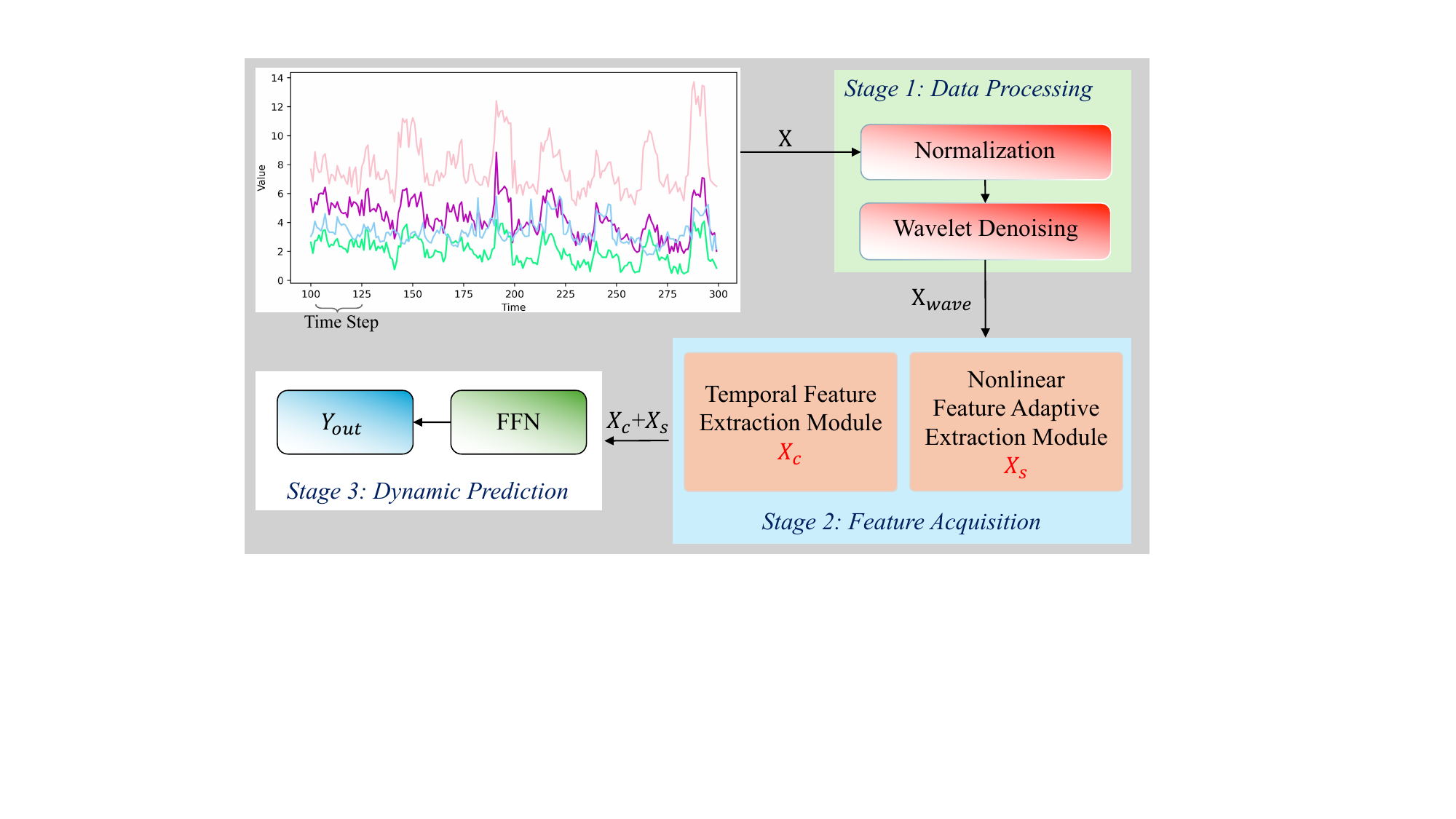}
\caption{Framework of the ACNet.}
\label{fig:modle}
\end{figure}

In this section, we will provide a detailed overview of the ACNet architecture. As depicted in Figure \ref{fig:modle}, the ACNet model is designed to effectively extract the time-domain pattern characteristics of time series and complex nonlinear information between different observations from multi-dimensional historical data to achieve accurate prediction of time series. The prediction process of the ACNet model mainly consists of the following three stages:

(1) Data processing: For a given multidimensional input dataset, first, standard normalization is applied to eliminate the influence of different resolutions, enhancing the robustness of the model. Subsequently, wavelet denoising is performed on the data to alleviate the interference of noise on model predictions.

(2) Feature acquisition: Fully extracting and utilizing the correlation information and nonlinear features between different time observations in a time series is key to accurately predicting TSF. We employ a temporal feature extraction module and a nonlinear feature adaptive extraction module to extract explicit correlated features hidden in the raw samples, which can significantly enhance prediction accuracy.

(3) Dynamic prediction: As time progresses, the distribution of time series data in real-world scenarios is likely to change, which may lead to decreased predictive accuracy of the model. Therefore, to maintain the model's performance, it is necessary to recalibrate the model parameters when there is a decline in predictive accuracy. This adjustment helps adapt to the features of new samples, ensuring the model's effectiveness in dynamic forecasting tasks.

\subsection{Problem statement}
In a rolling prediction setup with a fixed-size window, we consider an input sequence denoted as $X=\left \{ x_{1},\dots ,x_{i},\dots, x_{m} \right \}\in \mathbb{R} ^{M\times N} $, where $x_{i} =\left \{ x_{i}^{1},\dots,x_{i}^{j},\dots, x_{i}^{N } \right \} $, and $x_{i}^{j}$ represents the value of variable $j$ at the $i^{th}$ time point. The objective of long-term prediction for multivariate time series is to forecast $x_{m+L_{y}} =\left \{x_{m+1}, \cdots,x_{m+L_{y}}\right \}$, where the output length $L_{y}$ represents an extended future time period.

\subsection{Data processing}
To enhance the robustness of the model, we normalize the data and convert it into time series windows for training and testing. We employ a normalization method to adjust the time series data to a unified scale, expressed as
\begin{eqnarray}\label{formula:normalize}
 F(i)=\frac{x_{i}-\mu  }{\sigma } \quad, 
\end{eqnarray}
where $\mu$ and $\sigma$ are the mode-wise mean and variance vectors in the training time series. 

The accuracy of the model is constrained by the quality of the dataset, which directly impacts the model's predictive outcomes. To enhance the data quality of training samples and mitigate the influence of noisy data on model training, we employ a compromise method between soft and hard thresholds \cite{WOS:000867379900001}. This method involves filtering based on calculating the optimal denoising threshold, aiming to achieve effective noise reduction.

We assume the presence of a set of data affected by Gaussian white noise, which can be expressed using the following formula:
\begin{eqnarray}
	f_{i} =\theta_{i} +\varepsilon e_{i} ,
\end{eqnarray}
where $i$ is the $i^{th}$ sample, $i=1,2,\dots, m$, $e_{i}$ is white noise, $\theta_{i}$ is the data after denoising, and $\varepsilon$ is the level of noise.

The following is a detailed description of the wavelet transform thresholding denoising process:

First, perform an orthogonal wavelet transform and select $X$ historical data samples as the input for discrete wavelet transformation. Then, decompose the data into wavelet coefficients up to the $j^{th}$ layer. The wavelet decomposition coefficients $O_{j,k}$ for each layer can be calculated using the following formula:
\begin{eqnarray}
	O_{j,k}=\left \langle f,\Psi _{j,k}  \right \rangle =\int_{-\infty}^{+\infty}f(x)\Psi _{j,k} (t)dt,  
\end{eqnarray}
where $f$ represents the input sequence.

Next, the decomposed wavelet coefficients are subjected to thresholding, where each coefficient is compared with a predefined threshold to obtain the estimated wavelet coefficients. The soft-hard threshold compromise denoising method is expressed as 
\begin{eqnarray}
 \widehat{O_{j,k} } = \begin{cases}
  sgn\left ( O_{j,k}  \right ) \left ( \left | O_{j,k}  \right | -a\gamma  \right ) ,   \quad\quad \left | O_{j,k}  \right | \ge \gamma ,
  \\
  0, \quad \quad \left | O_{j,k}  \right | <  \gamma,
  \end{cases}
 \label{step}
\end{eqnarray}
where $O_{j,k}$ represents the wavelet coefficients obtained from signal decomposition, $\widehat{O_{j,k} }$ represents the wavelet coefficients obtained using the compromise threshold method, $\gamma$ represents the threshold value of wavelet denoising, $0\le a\le 1$ can obtain a better denoising effect.

Finally, wavelet reconstruction is performed on the estimated wavelet coefficients by using inverse wavelet transform to obtain denoised samples.
\begin{eqnarray}\label{formula:gauss}
	\widetilde{X(t)} =\sum_{j=-\infty }^{+\infty } \sum_{k=-\infty }^{+\infty }\widehat{O_{j,k} } \Psi _{j,k} .
\end{eqnarray}

\subsection{Feature acquisition}
\begin{figure}[!htbp]
	\centering
	\subfigure[Feature Acquisition module.]{
		\includegraphics[width=0.53\columnwidth]{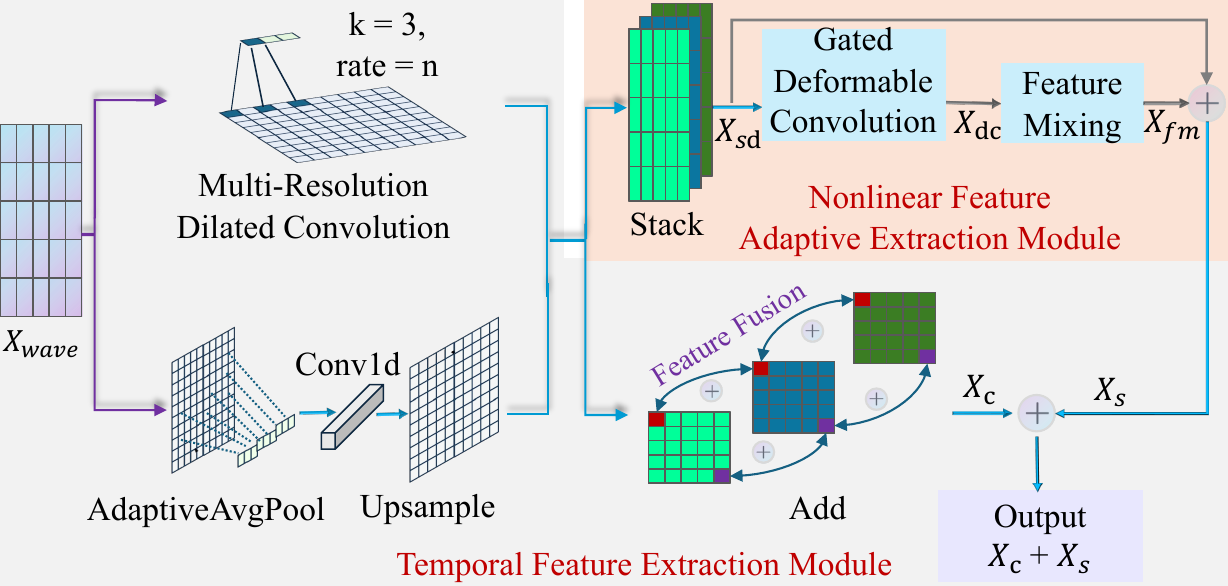}
	}
	\subfigure[Nonlinear Feature Adaptive Extraction Module.]{
		\includegraphics[width=6.6 cm]{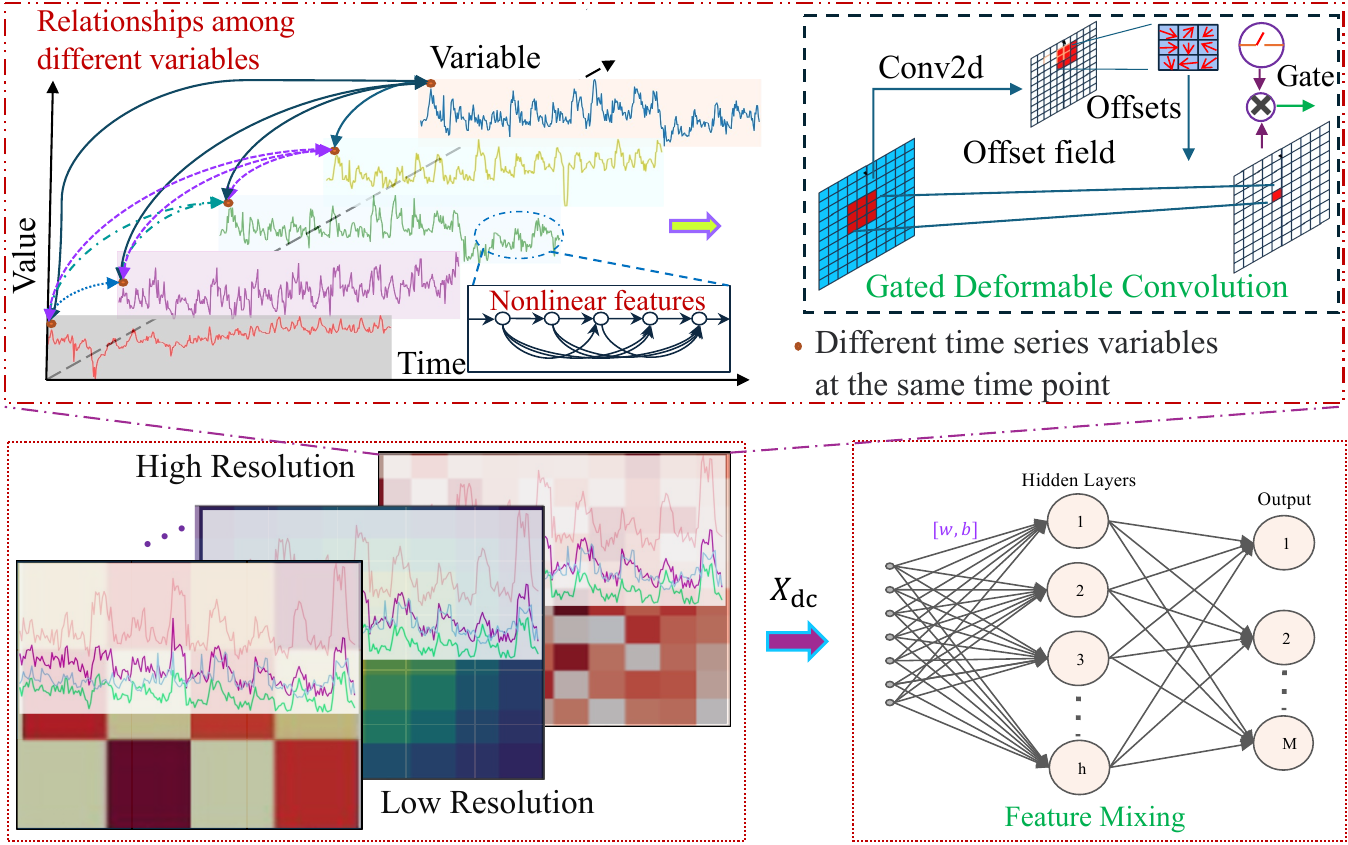}
	}
	\caption{Temporal-Nonlinear feature extraction module.}
	\label{fig:ATCNet}
\end{figure}

To comprehensively and effectively extract features from time series, 
we designed the feature acquisition module, which consists of two components: the temporal feature extraction module and the nonlinear feature adaptive extraction module, as shown in Figure \ref{fig:ATCNet} (a). 
The former is utilized to capture temporal dependency information of local and global patterns within the time series (as shown in the $Temporal Feature Extraction Module$ in Figure \ref{fig:ATCNet} (a)), while the latter is employed to effectively extract nonlinear information between different time observations and correlations among different variables at various resolutions in the time series. (as shown in the $Nonlinear Feature Adaptive Extraction Module$ in Figure \ref{fig:ATCNet} (a)).

\textbf{Temporal Feature Extraction Module }
Inspired by the concept of SPP \cite{Kaiming2015Spatial}, the temporal feature extraction module employs one-dimensional (1D) convolutions with varying dilation factors to capture features and patterns at different resolutions within the input multivariate time series data. This approach equips the model with the capability to discern local feature variations: with small dilation factors, the model convolves within a small receptive field, facilitating more precise identification of local patterns and structures, thus aiding in enhancing the network's comprehension of complex patterns. Conversely, with large dilation factors, the model employs larger convolutional kernels to capture features within a broader receptive field, thereby facilitating the identification of local contextual relationships within the input sequence. 

Additionally, by aggregating feature and pattern information extracted at different resolutions at the same location, it ensures positional invariance of the data during the processing. 
To encapsulate global trends and patterns, our model incorporates adaptive average pooling operations. This technique is instrumental in distilling the overarching trends and patterns from the time series data, thus enriching the model’s comprehensive grasp of the temporal dynamics presented.

When data flows through this module, it is mapped into two tasks: (1) Local feature extraction. By selecting different dilation factors for multi-resolution dilated convolution, the receptive field of the convolutional kernel can be expanded without losing resolution. This module utilizes convolutions of different sizes to learn patterns and local contextual information at different resolutions. (2) Global feature extraction. Introducing global adaptive average pooling effectively captures global information from the entire feature map, aiding the model in better understanding the global patterns and structural information of the entire input sequence.
In summary, by overlaying features of different resolutions at the same location, this not only enhances the model's translational invariance but also improves the model's generalization capability through the integration of complementary information.
Furthermore, through the designed weight sharing mechanism, the same set of weights is reused across inputs of different resolutions. This not only reduces the number of parameters in the model but also helps to enhance the model's efficiency.

The calculation formula for specific operational steps is as follows:

(1) Given a time series $X_{wave}$ with a length of $L$, a convolution kernel $w$ with a length of $k$, and a dilation factor $d$ ( by default, $d=1$), the formula for calculating the 1D dilated convolution is as follows:
\begin{eqnarray}
	Conv1d =\sum_{i=1}^{k}w_{i} x(t+ d\times i),
\end{eqnarray}
where $Conv1d$ represents the output value at time $t$, $x(t)$ represents the $t$-th element in the input sequence, $w_i$ represents the $i$th element of the convolution kernel $w$. The calculation formula indicates that at each time $t$, the output value $Conv1d$ is obtained by weighting and summing the local area of $x$.

(2) Temporal feature extraction. 
\begin{equation}
\begin{split}
	&\chi _{dilated}=Relu(LN(Conv1d(\chi_{i}^{l}, d=n))), n\in R, \\
	&\chi _{avg\_pool}=Upsample(Conv1d(AvgPool(\chi_{i}^{l}))).
	\label{eq_bns3}
\end{split}
\end{equation}

In the equations, $\chi _{dilated}$ is the result of processing the original data $\chi_{i}^{l} $ through convolution, linear normalization ($LN$) and ReLU activation function processing. To balance performance and efficiency, we set the dilated rate $n=\{1, 2, 5 \}$ as suggested in \cite{hybrid}. $\chi _{avg\_pool}$ is the result of adaptive average pooling after up sampling.

Through the aforementioned operations, we capture the local features and pattern information of the time series at different resolutions in $\chi _{dilated}$ and the global information of the time series in $\chi _{avg\_pool}$. Then, at the corresponding positions, features from different resolutions at the same time point are fused to obtain the final representation $X_{c}$ of cross-scale temporal feature information. This design allows the model to obtain contextual information from different temporal scales, providing more perspectives for inference and prediction while ensuring robustness and generalization.
\begin{eqnarray}
	X_{c} = Add(\chi _{dilated}, \chi _{avg\_pool}).
	\label{eq_bns4}
\end{eqnarray}

\textbf{Nonlinear Feature Adaptive Extraction Module}
Figure \ref{Nonlinear} illustrates that dependencies between different time observations in time series data may exhibit nonlinear and non-uniform distributions. 
To extract nonlinear features from complex time series, we designed a gated deformable convolution (GDC) module. Specifically, we added a gating mechanism to the deformable convolution, which further enhances the model's flexibility and capability, helping it to better capture nonlinear features in complex time series. By adjusting the positions of the convolutional kernels, deformable convolution enhances the model's ability to capture local nonlinear features, better handling irregular features and patterns, thereby improving its perception of local nonlinear patterns, as shown in Figure \ref{fig:ATCNet} (b).
This capability is particularly important for time series data with significant nonlinear and dynamic feature changes.



Specifically, we stack the extracted multi-resolution feature maps and the output maps from adaptive average pooling into a tensor $X_{sd}$, which serves as the input to the nonlinear feature adaptive extraction module. We employ GDC module to learn the nonlinear information between different observations and position offset information in feature graphs with different resolutions, focusing attention on regions with richer information through twist sampling networks. 

This operation enables the model to more accurately capture and understand the nonlinear features and relationships between different variables that vary between short-term and long-term in time series, while removing redundant feature information. 
The mathematical expression of the nonlinear feature extraction module is as follows:
\begin{equation}
DefConv(p_{0}) =  {\textstyle \sum_{p_{n}\in R}^{}} \mathcal{W} (p_{n})\cdot x(p_{0}+p_{n}+\triangle p_{n}) ,
\end{equation}
where $DefConv(p_{0})$ is the value of the output feature map at location $p_{0}$. $x(\cdot)$ represents the input feature map. $\mathcal{W} (p_{n})$ denotes the weights of the convolutional kernel. $\triangle p_{n}$ is the learned offset for each $p_{n}$, differing for each point, allowing the convolutional kernel to adaptively adjust its sampling positions on the input feature map. For more detailed theoretical explanations regarding deformable convolutions, please refer to reference \cite{Dai_Qi_Xiong_Li_Zhang_Hu_Wei_2017}.
The nonlinear feature adaptive extraction is described by the following equations:
\begin{equation}
\begin{split}
&X_{sd}=Stack(\chi _{dilated}, \chi _{avg\_pool}),\\
&Gate = Sigmoid(Conv2D(X_{sd})), \\
&X_{dc} = Gate * DefConv(X_{sd}),
\label{eq_def}
\end{split}
\end{equation}
where $Conv2D$ denotes a two-dimensional convolution operation, while $sigmoid$ represents the activation function.

Additionally, we implemented a feature mixing layer to integrate the lower-level fine-grained time series information with the higher-level coarse-grained time series information, aiming to merge the correlation information between features at different resolutions. Finally, the use of residual connections allows important feature information to propagate between network layers, preventing information loss. The process is described as follows:
\begin{equation}
\begin{split}
& X_{fm} = Linear(X_{dc}),\\
& X_{s} =  X_{sd} +  X_{fm} .  
\label{eq_bns5}
\end{split}
\end{equation}

In the end, the correlations between different dimensions in the time series are captured in $X_{s}$.

\subsection{Dynamic prediction}


\textbf{Model prediction.} In the downstream prediction tasks of ACNet, we employ a simple feedforward neural network (FFN) architecture. To address potential underfitting issues in regression tasks and improve the computational efficiency of the model, we utilize the Moore-Penrose pseudo-inverse matrix \cite{vasquez2023review} to compute the parameters of the model network quickly. The mathematical expression for the model prediction task is as follows:

\begin{equation}
    H \left(G \right)=\left [ h_{1}\left ( x_{1} \right ), h_{2}\left ( x_{2} \right ),\dots ,h_{L}\left ( x_{L} \right )  \right ],
\end{equation}
where $G=X_{c}+X_{s}$, $L$ is the number of hidden layer nodes, $h_{i}\left ( x_{i} \right )=sigmoid\left ( w_{i}x_{i}+b_{i}  \right ) $ represents the output of the $i${-}th hidden layer node, and $H\left(G \right)$ is the output matrix of the hidden layer. The weight matrix $\beta$ of the output layer is calculated by the least squares method:
\begin{equation}
    \beta = \left (H\left(G \right) ^{T}H\left(G \right)  \right ) ^{-1} H\left(G \right)^{T}Y ,
\end{equation}
where $Y$ is the target matrix. Ultimately, the output $y_{i}$ predicted by the model is
\begin{equation}
	y_{i} =\sum_{j=1}^{L} \beta _{j} h_{j} \left ( x_{i}  \right ) =H\left(G \right)\beta.
\end{equation}


\textbf{Model dynamic updates.} As time progresses, the data distribution in real-world applications may change, resulting in a significant deterioration of the model's predictive performance.
To address this issue, we extend the model training process to a dynamic update stage. We assume that when the model's predictive performance decreases by 5\%, we adopt newly collected samples to retrain the ACNet model. It is noteworthy that during this process, we maintain a fixed training set size, which means that when introducing new data, an equal amount of the earliest historical data is removed. This strategy ensures the speed of model training and reduces the required time. 

Algorithm \ref{alg:online} outlines the details of the dynamic prediction process.

\renewcommand{\algorithmicrequire}{\textbf{Input:}}
\renewcommand{\algorithmicensure}{\textbf{Output:}}
\begin{algorithm}[htb] 
	\caption{: Dynamic prediction.}
	\label{alg:online}
	\begin{algorithmic}[1]
	\REQUIRE ~~\\$X=\left \{ x_{1},\dots ,x_{i}, \dots,x_{m} \right \} $: $X$ denotes the training dataset, where $x_{i}$ represents the $N$ variable factors at time $i$.\\
	\STATE \textbf{Phase \uppercase\expandafter{\romannumeral1}.} Training the proposed ACNet model.
	\STATE \textbf{Randomly generated:} The weight $\omega$ and the bias vector $b$ between the input layer and the hidden layer in FFN.
	\STATE Data processing $X_{wave}$ is obtained based on formulas ~\eqref{formula:normalize} to ~\eqref{formula:gauss}.
	\STATE Extract temporal correlation $X_{c}$ based on formulas ~\eqref{eq_bns3} to ~\eqref{eq_bns4}
	\STATE Extract inter-variable correlation $X_{s}$ based on formulas  ~\eqref{eq_def} to ~\eqref{eq_bns5}
	\STATE Input $X_{c}+X_{s}$ into the FFN and calculate $\beta$.
	\STATE \textbf{Phase \uppercase\expandafter{\romannumeral2}}.Dynamically forecasting the values of a time series.
	\STATE Input the obtained new samples $ x_{j}$ and training data set $X$ into the trained ACNet model. If the MSE value exceeds the threshold, update $\beta$.
	\ENSURE Predicting the trend of changes over a future time period.
	\end{algorithmic}
	
\end{algorithm}

\section{Experiments}\label{sec:sample4}

In this section, we will delve into the effectiveness and efficiency of the ACNet network. Through rigorous experiments on long-term and short-term forecasting using twelve real-world datasets and comparing against fifteen SOTA models, our goal is to address the following research questions:
\begin{itemize}
\item 
RQ1 (Effectiveness): Can ACNet outshine the present SOTA baseline models when applied to real datasets (Sections \ref{main} - \ref{nonlinear})?
\item 
RQ2 (Efficiency): Does ACNet exhibit superior performance in terms of resource utilization compared to the current baseline models (Sections \ref{overhead})?
\item 
RQ3 (Ablation): To what extent do the distinct components of ACNet influence its overall performance in time series forecasting tasks (Sections \ref{ab})?
\end{itemize}

\subsection{Datasets}

We extensively evaluated the proposed ACNet model on twelve benchmark datasets spanning four real-world fields: energy, finance, medical, and traffic. Table \ref{data} summarizes the key features of these datasets.
\begin{itemize}

\item \textbf{ETT\footnote{ETT dataset was acquired at https:// github.com/zhouhaoyi/ETDataset}:} The dataset records continuous operation data of power resources over a period of two years, including seven indicators such as oil temperature and load. 
\item \textbf{Electricity\footnote{https://archieve.ics.uci.deu/ml/datasets/ElectricityLoadDiagrams 20112014}:} The dataset comprises hourly electricity consumption data for 321 customers over two years.
\item \textbf{Traffic\footnote{http://pems.dot.ca.gov}:} The dataset consists of measurements collected every hour from 862 sensors located on the highways in the San Francisco Bay Area.
\item \textbf{Exchange\footnote{https://github.com/laiguo-kun/multivariate-time-series-data}:} The dataset records the daily exchange rates of eight different countries over a period of 26 years.
\item \textbf{ILI \footnote{https://gis.cdc.gov/grasp/fluview/fluportaldashboard.html}:} The dataset collects seven indicators, including the proportion of influenza patients and the total number of patients, from the Centers for Disease Control and Prevention in the United States on a weekly basis.

\item \textbf{PEMS\cite{chen2001freeway}:} This dataset records traffic network data collected every 5 minutes on California highways, including metrics such as traffic flow, speed, and occupancy.

\end{itemize}

\begin{table}[h]
\centering
\caption{The key features of the twelve time series datasets.}
\label{data}
\begin{tabular}{l|lllllllll} 
\hline
Tasks                                                                             & Datasets    &  & Dim &  & Timesteps &  & Granularity &  & Information  \\ 
\hline
\multirow{6}{*}{\begin{tabular}[c]{@{}l@{}}Long-term~\\Forecasting\end{tabular}}  & ETTh1/h2    &  & 7        &  & 17420     &  & 1 hour      &  & Electricity  \\
                                                                                  & ETTm1/m2    &  & 7        &  & 69680     &  & 15 min      &  & Electricity  \\
                                                                                  & Electricity &  & 321      &  & 26304     &  & 1 hour      &  & Electricity  \\
                                                                                  & Traffic     &  & 862      &  & 52560     &  & 1 hour      &  & Traffic      \\
                                                                                  & Exchange    &  & 8        &  & 7558      &  & 1 day       &  & Financial    \\
                                                                                  & ILI         &  & 7        &  & 966       &  & 1 week      &  & Medical      \\ 
\hline
\multirow{4}{*}{\begin{tabular}[c]{@{}l@{}}Short-term~\\Forecasting\end{tabular}} & PEMS03      &  & 358      &  & 26208     &  & 5 min       &  & Traffic      \\
                                                                                  & PEMS04      &  & 307      &  & 16992     &  & 5 min       &  & Traffic      \\
                                                                                  & PEMS07      &  & 883      &  & 28224     &  & 5 min       &  & Traffic      \\
                                                                                  & PEMS08      &  & 170      &  & 17856     &  & 5 min       &  & Traffic      \\
\hline
\end{tabular}
\end{table}

\subsection{Baselines}
 
 We selected 15 representative SOTA models that have demonstrated outstanding performance in the field of TSF as benchmarks: Patch-based TSMixer \cite{103}, PatchTST \cite{nie2023a}, and PDF \cite{dai2023periodicity}; MLP-based TimeMixer \cite{wang2024timemixer}, FITS \cite{xu2024fits}, and DLinear \cite{zeng2023transformers}; Transformer-based architectures iTransformer \cite{liu2023itransformer}, Crossformer \cite{hassanin2022crossformer}, ETSformer \cite{woo2022etsformer}, FEDformer \cite{zhou2022fedformer}, and Non-Stationary Transformer \cite{liu2022non}; and CNN-based models ConvTimeNet \cite{cheng2024convtimenet}, TSLANet \cite{eldele2024tslanet}, MICN \cite{wang2022micn}, and TimesNet \cite{wu2022timesnet}.

\subsection{Implementation details}
Our approach is trained with an initial learning rate of $1 \times 10^{-3}$, the convolution dilated rate $d$ for multi-resolution dilated convolution was set to $\left \{ 1, 2, 5 \right \}$. In long-term forecasting, the input sequence length for the ILI dataset is $L=36$, with prediction sequence lengths of $\left \{24, 36, 48, 60 \right \}$. For other datasets, the input sequence length is $L=96$, with prediction sequence lengths of $\left \{96, 192, 336, 720\right \}$. In short-term forecasting, the input sequence length for the PEMS datasets is $L=96$, with a prediction sequence length of $12$.
To verify the robustness of our results, we conducted long-term forecasting by training the ACNet model with three different random seeds. For each seed, we calculated the mean squared error (MSE) and mean absolute error (MAE). The average results are presented in Table \ref{result}. For short-term forecasting, we employed mean absolute error (MAE), mean absolute percentage error (MAPE), and root mean squared error (RMSE) as evaluation metrics. Additionally, we assessed the model's efficiency based on floating point operations (FLOPs), number of parameters, training time, and inference time. Experiments are implemented using PyTorch on a single NVIDIA GeForce RTX 3090 24GB GPU, Intel(R) Core(TM) i7-10700k CPU and 32GB RAM.


\subsection{Main results and analysis} \label{main}



\textbf{Long-term forecasting.}
In the field of long-term forecasting, the ACNet framework demonstrates excellent performance on almost all datasets. Through analysis of the result data in Table \ref{result}, the following conclusions can be drawn: 
\begin{itemize}
    \item The ACNet model significantly improves inference performance on nearly all datasets. As the forecasting horizon increases, the prediction error of ACNet shows a gradually stable upward trend, indicating its ability to maintain high long-term robustness in practical applications. 
\item Compared to the current SOTA forecasting model ConvTimeNet, ACNet exhibits notable improvements across various datasets. Specifically, ACNet gives 47.9\% $\left (0.319 \to 0.166\right)$ MSE reduction in ETTm1, 36.9\% $\left (0.211 \to 0.133\right)$ in ETTm2, 42\% $\left (0.395 \to 0.229\right)$ in ETTh1, 58\% $\left (0.436 \to 0.183\right)$ in ETTh2, 32.1\% $\left (0.202 \to \notag \right.\\ \left. 0.137\right)$ in Electricity, 39.8\% $\left (0.402 \to 0.242\right)$ in Traffic, 65.7\% $\left (0.347 \to 0.119\right)$ in exchange and 82.9\% $\left (1.866 \to 0.319\right)$ in ILI. Overall, ACNet yields a 63.4\% averaged MSE reduction among above settings.

Notably, ACNet has achieved great success on the Exchange and ILI datasets, which contain a large number of nonlinear features. This demonstrates the effectiveness of the ACNet model in handling nonlinear features.
Notably, ACNet achieved significant success on the Exchange and ILI datasets, which contain a large number of nonlinear features. Compared to the ConvTimeNet model, which uses depthwise separable convolutions to extract time series features, ACNet's extraction of nonlinear features and inter-variable correlations at different resolution scales proves to be more advantageous for TSF.


\item ACNet outperforms both the iTransformer and TSMixer models significantly. Compared to iTransformer, ACNet reduces the MSE across all datasets by 66.8\% $\left (0.576 \to 0.191\right)$. Similarly, compared to TSMixer, ACNet reduces the MSE across all datasets by 49.6\% $\left (0.379 \to 0.191\right)$. iTransformer and TSMixer both emphasize feature extraction between different variables in time series. While they perform better than models such as PatchTST, which focuses solely on extracting temporal information, they may be affected by strong correlations between different variables, leading to information redundancy and decreased prediction accuracy. ACNet uses deformable convolution to extract the correlation between different variables in the time series and uses a flexible warp network to focus on important features, remove redundant features, and improve the prediction performance of the model.

\item 
The ACNet model achieves significantly better results than the SOTA MLP-based models. 
ACNet has an average MSE decrease of 67.5\% $\left (0.587 \to 0.191\right)$ on all datasets in TimeMixer and 49.6\% $\left (0.379 \to 0.191\right)$ in TSMixer. 
We analyzed that this may be because MLP-based models need to extract more feature information from complex time series to fit future trends. Due to the comprehensive feature extraction capability of the model, ACNet can not only extract and utilize local contextual information and global correlation information of time series but also extract information between variables at different resolutions as a complement to TSF, which improves the prediction accuracy.

\end{itemize}

In summary, the excellent performance of ACNet in handling long time series may be attributed to its comprehensive consideration of local information, global information, and interdependencies between different variables in the time series, aspects that many current TSF models do not take into account. This allows ACNet to excel in handling various real-world time series prediction tasks.


\begin{table*}[h]
\centering
\caption{Long-term multivariate forecasting results with different prediction lengths. The best results are in bold numbers. Avg represents the average value across the four prediction lengths.}\label{result}
\resizebox{\columnwidth}{!}{
\begin{tabular}{c|c|cc|cc|cc|cc|cc|cc|cc|cc|cc|cc|cc|cc|cc|cc|cc} 
\hline
\multicolumn{2}{c|}{Models}        & \multicolumn{2}{c|}{\begin{tabular}[c]{@{}c@{}}ACNet\\(Ours)\end{tabular}} & \multicolumn{2}{c|}{\begin{tabular}[c]{@{}c@{}}ConvTimeNet\\(2024)\end{tabular}} & \multicolumn{2}{c|}{\begin{tabular}[c]{@{}c@{}}TimeMixer\\(2024)\end{tabular}} & \multicolumn{2}{c|}{\begin{tabular}[c]{@{}c@{}}iTransformer\\(2024)\end{tabular}} & \multicolumn{2}{c|}{\begin{tabular}[c]{@{}c@{}}FITS\\(2024)\end{tabular}} & \multicolumn{2}{c|}{\begin{tabular}[c]{@{}c@{}}PDF\\(2024)\end{tabular}} & \multicolumn{2}{c|}{\begin{tabular}[c]{@{}c@{}}TSMixer\\(2023)\end{tabular}} & \multicolumn{2}{c|}{\begin{tabular}[c]{@{}c@{}}PatchTST\\(2023)\end{tabular}} & \multicolumn{2}{c|}{\begin{tabular}[c]{@{}c@{}}MICN\\(2023)\end{tabular}} & \multicolumn{2}{c|}{\begin{tabular}[c]{@{}c@{}}TimesNet\\(2023)\end{tabular}} & \multicolumn{2}{c|}{\begin{tabular}[c]{@{}c@{}}Crossformer\\(2023)\end{tabular}} & \multicolumn{2}{c|}{\begin{tabular}[c]{@{}c@{}}DLinear\\(2023)\end{tabular}} & \multicolumn{2}{c|}{\begin{tabular}[c]{@{}c@{}}ETSformer\\(2022)\end{tabular}} & \multicolumn{2}{c|}{\begin{tabular}[c]{@{}c@{}}FEDformer\\(2022)\end{tabular}} & \multicolumn{2}{c}{\begin{tabular}[c]{@{}c@{}}Non-Sta\\(2022)\end{tabular}}  \\ 
\hline
\multicolumn{2}{c|}{Metric}        & MSE             & MAE                                                        & MSE    & MAE                                                                     & MSE    & MAE                                                                   & MSE    & MAE                                                                      & MSE    & MAE                                                              & MSE    & MAE                                                             & MSE             & MAE                                                        & MSE             & MAE                                                         & MSE    & MAE                                                              & MSE    & MAE                                                                  & MSE    & MAE                                                                     & MSE    & MAE                                                                 & MSE    & MAE                                                                   & MSE    & MAE                                                                   & MSE    & MAE                                                                 \\ 
\hline
\multirow{5}{*}{ETTm1}       & 96  & \textbf{0.142~} & \textbf{0.263~}                                            & 0.264~ & 0.330~                                                                  & 0.320~ & 0.357~                                                                & 0.334~ & 0.368~                                                                   & 0.351~ & 0.370~                                                           & 0.320~ & 0.351~                                                          & 0.319~          & 0.358~                                                     & 0.324~          & 0.361~                                                      & 0.316~ & 0.362~                                                           & 0.338~ & 0.375~                                                               & 0.428~ & 0.444~                                                                  & 0.345~ & 0.372~                                                              & 0.375~ & 0.398~                                                                & 0.379~ & 0.419~                                                                & 0.386~ & 0.398~                                                              \\
                             & 192 & \textbf{0.162~} & \textbf{0.282~}                                            & 0.316~ & 0.368~                                                                  & 0.361~ & 0.381~                                                                & 0.377~ & 0.391~                                                                   & 0.392~ & 0.393~                                                           & 0.375~ & 0.376~                                                          & 0.369~          & 0.384~                                                     & 0.362~          & 0.383~                                                      & 0.363~ & 0.390~                                                           & 0.374~ & 0.387~                                                               & 0.445~ & 0.468~                                                                  & 0.380~ & 0.389~                                                              & 0.408~ & 0.410~                                                                & 0.426~ & 0.441~                                                                & 0.459~ & 0.444~                                                              \\
                             & 336 & \textbf{0.174~} & \textbf{0.295~}                                            & 0.315~ & 0.378~                                                                  & 0.390~ & 0.404~                                                                & 0.426~ & 0.420~                                                                   & 0.424~ & 0.413~                                                           & 0.411~ & 0.399~                                                          & 0.402~          & 0.406~                                                     & 0.390~          & 0.402~                                                      & 0.408~ & 0.426~                                                           & 0.410~ & 0.411~                                                               & 0.533~ & 0.519~                                                                  & 0.413~ & 0.413~                                                              & 0.435~ & 0.428~                                                                & 0.445~ & 0.459~                                                                & 0.495~ & 0.464~                                                              \\
                             & 720 & \textbf{0.184~} & \textbf{0.304~}                                            & 0.382~ & 0.425~                                                                  & 0.454~ & 0.441~                                                                & 0.491~ & 0.459~                                                                   & 0.485~ & 0.448~                                                           & 0.464~ & 0.431~                                                          & 0.464~          & 0.442~                                                     & 0.461~          & 0.438~                                                      & 0.481~ & 0.476~                                                           & 0.478~ & 0.450~                                                               & 0.728~ & 0.655~                                                                  & 0.474~ & 0.453~                                                              & 0.499~ & 0.462~                                                                & 0.543~ & 0.490~                                                                & 0.585~ & 0.516~                                                              \\ 
\cline{2-32}
                             & Avg & \textbf{0.166~} & \textbf{0.286~}                                            & 0.319~ & 0.375~                                                                  & 0.381~ & 0.395~                                                                & 0.407~ & 0.410~                                                                   & 0.413~ & 0.406~                                                           & 0.393~ & 0.389~                                                          & 0.389~          & 0.398~                                                     & 0.384~          & 0.396~                                                      & 0.392~ & 0.414~                                                           & 0.400~ & 0.406~                                                               & 0.534~ & 0.522~                                                                  & 0.403~ & 0.407~                                                              & 0.429~ & 0.425~                                                                & 0.448~ & 0.452~                                                                & 0.481~ & 0.456~                                                              \\ 
\hline
\multirow{5}{*}{ETTm2}       & 96  & \textbf{0.111~} & \textbf{0.219~}                                            & 0.183~ & 0.265~                                                                  & 0.175~ & 0.258~                                                                & 0.180~ & 0.264~                                                                   & 0.181~ & 0.264~                                                           & 0.181~ & 0.264~                                                          & 0.175~          & 0.258~                                                     & 0.177~          & 0.260~                                                      & 0.179~ & 0.275~                                                           & 0.187~ & 0.267~                                                               & 0.197~ & 0.321~                                                                  & 0.193~ & 0.292~                                                              & 0.189~ & 0.280~                                                                & 0.203~ & 0.287~                                                                & 0.192~ & 0.274~                                                              \\
                             & 192 & \textbf{0.129~} & \textbf{0.236~}                                            & 0.248~ & 0.305~                                                                  & 0.237~ & 0.299~                                                                & 0.250~ & 0.309~                                                                   & 0.246~ & 0.304~                                                           & 0.242~ & 0.302~                                                          & 0.240~          & 0.230~                                                     & 0.248~          & 0.306~                                                      & 0.307~ & 0.376~                                                           & 0.249~ & 0.309~                                                               & 0.326~ & 0.375~                                                                  & 0.284~ & 0.362~                                                              & 0.253~ & 0.319~                                                                & 0.269~ & 0.328~                                                                & 0.280~ & 0.339~                                                              \\
                             & 336 & \textbf{0.140~} & \textbf{0.247~}                                            & 0.311~ & 0.345~                                                                  & 0.298~ & 0.340~                                                                & 0.311~ & 0.348~                                                                   & 0.306~ & 0.341~                                                           & 0.302~ & 0.341~                                                          & 0.301~          & 0.338~                                                     & 0.304~          & 0.342~                                                      & 0.325~ & 0.388~                                                           & 0.321~ & 0.351~                                                               & 0.372~ & 0.421~                                                                  & 0.369~ & 0.427~                                                              & 0.314~ & 0.357~                                                                & 0.325~ & 0.366~                                                                & 0.334~ & 0.361~                                                              \\
                             & 720 & \textbf{0.152~} & \textbf{0.255~}                                            & 0.101~ & 0.396~                                                                  & 0.391~ & 0.396~                                                                & 0.412~ & 0.407~                                                                   & 0.407~ & 0.397~                                                           & 0.403~ & 0.396~                                                          & 0.404~          & 0.398~                                                     & 0.403~          & 0.397~                                                      & 0.502~ & 0.490~                                                           & 0.408~ & 0.403~                                                               & 0.410~ & 0.448~                                                                  & 0.554~ & 0.522~                                                              & 0.414~ & 0.413~                                                                & 0.421~ & 0.415~                                                                & 0.417~ & 0.413~                                                              \\ 
\cline{2-32}
                             & Avg & \textbf{0.133~} & \textbf{0.239~}                                            & 0.211~ & 0.328~                                                                  & 0.275~ & 0.323~                                                                & 0.288~ & 0.332~                                                                   & 0.285~ & 0.327~                                                           & 0.282~ & 0.326~                                                          & 0.280~          & 0.306~                                                     & 0.283~          & 0.326~                                                      & 0.328~ & 0.382~                                                           & 0.291~ & 0.333~                                                               & 0.326~ & 0.391~                                                                  & 0.350~ & 0.401~                                                              & 0.293~ & 0.342~                                                                & 0.305~ & 0.349~                                                                & 0.306~ & 0.347~                                                              \\ 
\hline
\multirow{5}{*}{ETTh1}       & 96  & \textbf{0.203~} & \textbf{0.326~}                                            & 0.338~ & 0.368~                                                                  & 0.375~ & 0.400~                                                                & 0.386~ & 0.405~                                                                   & 0.381~ & 0.391~                                                           & 0.369~ & 0.387~                                                          & 0.381~          & 0.391~                                                     & 0.394~          & 0.408~                                                      & 0.398~ & 0.427~                                                           & 0.384~ & 0.402~                                                               & 0.429~ & 0.440~                                                                  & 0.386~ & 0.400~                                                              & 0.494~ & 0.479~                                                                & 0.376~ & 0.419~                                                                & 0.513~ & 0.491~                                                              \\
                             & 192 & \textbf{0.228~} & \multicolumn{1}{c}{\textbf{0.347~}}                        & 0.377~ & 0.385~                                                                  & 0.429~ & 0.421~                                                                & 0.441~ & 0.436~                                                                   & 0.434~ & 0.422~                                                           & 0.414~ & 0.419~                                                          & 0.433~          & 0.420~                                                     & 0.446~          & 0.438~                                                      & 0.430~ & 0.453~                                                           & 0.436~ & 0.429~                                                               & 0.494~ & 0.482~                                                                  & 0.437~ & 0.432~                                                              & 0.538~ & 0.504~                                                                & 0.420~ & 0.448~                                                                & 0.534~ & 0.504~                                                              \\
                             & 336 & \textbf{0.238~} & \textbf{0.355~}                                            & 0.400~ & 0.404~                                                                  & 0.484~ & 0.458~                                                                & 0.487~ & 0.458~                                                                   & 0.474~ & 0.446~                                                           & 0.451~ & 0.438~                                                          & 0.472~          & 0.441~                                                     & 0.485~          & 0.455~                                                      & 0.460~ & 0.460~                                                           & 0.491~ & 0.469~                                                               & 0.706~ & 0.625~                                                                  & 0.481~ & 0.459~                                                              & 0.574~ & 0.521~                                                                & 0.459~ & 0.465~                                                                & 0.588~ & 0.535~                                                              \\
                             & 720 & \textbf{0.245~} & \textbf{0.363~}                                            & 0.465~ & 0.439~                                                                  & 0.498~ & 0.482~                                                                & 0.503~ & 0.491~                                                                   & 0.464~ & 0.463~                                                           & 0.483~ & 0.475~                                                          & 0.485~          & 0.471~                                                     & 0.495~          & 0.474~                                                      & 0.491~ & 0.509~                                                           & 0.521~ & 0.500~                                                               & 0.750~ & 0.689~                                                                  & 0.519~ & 0.516~                                                              & 0.562~ & 0.535~                                                                & 0.506~ & 0.507~                                                                & 0.643~ & 0.616~                                                              \\ 
\cline{2-32}
                             & Avg & \textbf{0.229~} & \textbf{0.348~}                                            & 0.395~ & 0.399~                                                                  & 0.447~ & 0.440~                                                                & 0.454~ & 0.448~                                                                   & 0.438~ & 0.431~                                                           & 0.429~ & 0.430~                                                          & 0.443~          & 0.431~                                                     & 0.455~          & 0.444~                                                      & 0.445~ & 0.462~                                                           & 0.458~ & 0.450~                                                               & 0.595~ & 0.559~                                                                  & 0.456~ & 0.452~                                                              & 0.542~ & 0.510~                                                                & 0.440~ & 0.460~                                                                & 0.570~ & 0.537~                                                              \\ 
\hline
\multirow{5}{*}{ETTh2}       & 96  & \textbf{0.147~} & \textbf{0.259~}                                            & 0.385~ & 0.396~                                                                  & 0.289~ & 0.341~                                                                & 0.297~ & 0.349~                                                                   & 0.290~ & 0.339~                                                           & 0.292~ & 0.341~                                                          & 0.289~          & 0.338~                                                     & 0.302~          & 0.348~                                                      & 0.332~ & 0.377~                                                           & 0.340~ & 0.374~                                                               & 0.632~ & 0.547~                                                                  & 0.333~ & 0.387~                                                              & 0.340~ & 0.391~                                                                & 0.358~ & 0.397~                                                                & 0.476~ & 0.458~                                                              \\
                             & 192 & \textbf{0.168~} & \textbf{0.279~}                                            & 0.438~ & 0.425~                                                                  & 0.372~ & 0.392~                                                                & 0.380~ & 0.400~                                                                   & 0.375~ & 0.388~                                                           & 0.377~ & 0.392~                                                          & 0.375~          & 0.391~                                                     & 0.388~          & 0.400~                                                      & 0.422~ & 0.441~                                                           & 0.402~ & 0.414~                                                               & 0.876~ & 0.663~                                                                  & 0.477~ & 0.476~                                                              & 0.430~ & 0.439~                                                                & 0.429~ & 0.439~                                                                & 0.512~ & 0.493~                                                              \\
                             & 336 & \textbf{0.192~} & \textbf{0.295~}                                            & 0.451~ & 0.437~                                                                  & 0.386~ & 0.414~                                                                & 0.428~ & 0.432~                                                                   & 0.414~ & 0.425~                                                           & 0.419~ & 0.429~                                                          & 0.425~          & 0.435~                                                     & 0.426~          & 0.433~                                                      & 0.447~ & 0.474~                                                           & 0.452~ & 0.452~                                                               & 0.924~ & 0.702~                                                                  & 0.594~ & 0.541~                                                              & 0.485~ & 0.479~                                                                & 0.496~ & 0.487~                                                                & 0.552~ & 0.551~                                                              \\
                             & 720 & \textbf{0.223~} & \textbf{0.310~}                                            & 0.470~ & 0.461~                                                                  & 0.412~ & 0.434~                                                                & 0.427~ & 0.445~                                                                   & 0.419~ & 0.437~                                                           & 0.424~ & 0.441~                                                          & 0.435~          & 0.449~                                                     & 0.431~          & 0.446~                                                      & 0.442~ & 0.467~                                                           & 0.462~ & 0.468~                                                               & 1.390~ & 0.863~                                                                  & 0.831~ & 0.657~                                                              & 0.500~ & 0.497~                                                                & 0.463~ & 0.474~                                                                & 0.562~ & 0.560~                                                              \\ 
\cline{2-32}
                             & Avg & \textbf{0.183~} & \textbf{0.286~}                                            & 0.436~ & 0.430~                                                                  & 0.364~ & 0.395~                                                                & 0.383~ & 0.407~                                                                   & 0.375~ & 0.397~                                                           & 0.378~ & 0.401~                                                          & 0.381~          & 0.403~                                                     & 0.387~          & 0.407~                                                      & 0.411~ & 0.440~                                                           & 0.414~ & 0.427~                                                               & 0.956~ & 0.694~                                                                  & 0.559~ & 0.515~                                                              & 0.439~ & 0.452~                                                                & 0.437~ & 0.449~                                                                & 0.526~ & 0.516~                                                              \\ 
\hline
\multirow{5}{*}{Electricity} & 96  & \textbf{0.131~} & \textbf{0.233~}                                            & 0.179~ & 0.263~                                                                  & 0.153~ & 0.247~                                                                & 0.148~ & 0.240~                                                                   & 0.293~ & 0.401~                                                           & 0.165~ & 0.248~                                                          & 0.146~          & 0.244~                                                     & 0.195~          & 0.285~                                                      & 0.164~ & 0.269~                                                           & 0.168~ & 0.272~                                                               & 0.254~ & 0.347~                                                                  & 0.197~ & 0.282~                                                              & 0.187~ & 0.304~                                                                & 0.193~ & 0.308~                                                                & 0.169~ & 0.273~                                                              \\
                             & 192 & \textbf{0.135~} & \textbf{0.239~}                                            & 0.185~ & 0.269~                                                                  & 0.166~ & 0.256~                                                                & 0.162~ & 0.253~                                                                   & 0.268~ & 0.378~                                                           & 0.181~ & 0.266~                                                          & 0.163~          & 0.259~                                                     & 0.199~          & 0.289~                                                      & 0.177~ & 0.285~                                                           & 0.184~ & 0.289~                                                               & 0.261~ & 0.353~                                                                  & 0.196~ & 0.285~                                                              & 0.199~ & 0.315~                                                                & 0.201~ & 0.315~                                                                & 0.182~ & 0.286~                                                              \\
                             & 336 & \textbf{0.138~} & \textbf{0.245~}                                            & 0.201~ & 0.285~                                                                  & 0.185~ & 0.277~                                                                & 0.178~ & 0.269~                                                                   & 0.355~ & 0.452~                                                           & 0.197~ & 0.282~                                                          & 0.180~          & 0.279~                                                     & 0.215~          & 0.305~                                                      & 0.193~ & 0.304~                                                           & 0.198~ & 0.300~                                                               & 0.273~ & 0.364~                                                                  & 0.209~ & 0.301~                                                              & 0.212~ & 0.329~                                                                & 0.214~ & 0.329~                                                                & 0.200~ & 0.304~                                                              \\
                             & 720 & \textbf{0.142~} & \textbf{0.250~}                                            & 0.242~ & 0.318~                                                                  & 0.225~ & 0.310~                                                                & 0.225~ & 0.317~                                                                   & 0.416~ & 0.498~                                                           & 0.238~ & 0.315~                                                          & 0.216~          & 0.307~                                                     & 0.256~          & 0.337~                                                      & 0.212~ & 0.321~                                                           & 0.220~ & 0.320~                                                               & 0.303~ & 0.388~                                                                  & 0.245~ & 0.333~                                                              & 0.233~ & 0.345~                                                                & 0.246~ & 0.355~                                                                & 0.222~ & 0.321~                                                              \\ 
\cline{2-32}
                             & Avg & \textbf{0.137~} & \textbf{0.242~}                                            & 0.202~ & 0.284~                                                                  & 0.182~ & 0.272~                                                                & 0.178~ & 0.270~                                                                   & 0.333~ & 0.432~                                                           & 0.195~ & 0.278~                                                          & 0.176~          & 0.272~                                                     & 0.216~          & 0.304~                                                      & 0.187~ & 0.295~                                                           & 0.193~ & 0.295~                                                               & 0.273~ & 0.363~                                                                  & 0.212~ & 0.300~                                                              & 0.208~ & 0.323~                                                                & 0.214~ & 0.327~                                                                & 0.193~ & 0.296~                                                              \\ 
\hline
\multirow{5}{*}{Traffic}     & 96  & \textbf{0.228~} & 0.286~                                                     & 0.376~ & \textbf{0.265~}                                                         & 0.462~ & 0.285~                                                                & 0.395~ & 0.268~                                                                   & 0.898~ & 0.572~                                                           & 0.463~ & 0.297~                                                          & 0.483~          & 0.321~                                                     & 0.544~          & 0.359~                                                      & 0.519~ & 0.309~                                                           & 0.593~ & 0.321~                                                               & 0.558~ & 0.320~                                                                  & 0.650~ & 0.396~                                                              & 0.607~ & 0.392~                                                                & 0.587~ & 0.366~                                                                & 0.612~ & 0.338~                                                              \\
                             & 192 & \textbf{0.239~} & 0.292~                                                     & 0.392~ & \textbf{0.271~}                                                         & 0.473~ & 0.296~                                                                & 0.417~ & 0.276~                                                                   & 0.763~ & 0.522~                                                           & 0.469~ & 0.298~                                                          & 0.490~          & 0.321~                                                     & 0.540~          & 0.354~                                                      & 0.537~ & 0.315~                                                           & 0.617~ & 0.336~                                                               & 0.572~ & 0.331~                                                                  & 0.598~ & 0.370~                                                              & 0.621~ & 0.399~                                                                & 0.604~ & 0.373~                                                                & 0.613~ & 0.340~                                                              \\
                             & 336 & \textbf{0.243~} & 0.300~                                                     & 0.405~ & \textbf{0.277~}                                                         & 0.498~ & 0.296~                                                                & 0.433~ & 0.283~                                                                   & 0.894~ & 0.608~                                                           & 0.484~ & 0.305~                                                          & 0.506~          & 0.330~                                                     & 0.551~          & 0.358~                                                      & 0.534~ & 0.313~                                                           & 0.629~ & 0.336~                                                               & 0.587~ & 0.342~                                                                  & 0.605~ & 0.373~                                                              & 0.622~ & 0.396~                                                                & 0.621~ & 0.383~                                                                & 0.618~ & 0.328~                                                              \\
                             & 720 & \textbf{0.256~} & 0.307~                                                     & 0.436~ & \textbf{0.294~}                                                         & 0.506~ & 0.313~                                                                & 0.467~ & 0.302~                                                                   & 1.019~ & 0.646~                                                           & -      & -                                                               & 0.538~          & 0.341~                                                     & 0.586~          & 0.375~                                                      & 0.577~ & 0.325~                                                           & 0.640~ & 0.350~                                                               & 0.652~ & 0.359~                                                                  & 0.645~ & 0.394~                                                              & 0.632~ & 0.396~                                                                & 0.626~ & 0.382~                                                                & 0.653~ & 0.355~                                                              \\ 
\cline{2-32}
                             & Avg & \textbf{0.242~} & 0.296~                                                     & 0.402~ & \textbf{0.277~}                                                         & 0.484~ & 0.297~                                                                & 0.428~ & 0.282~                                                                   & 0.894~ & 0.587~                                                           & 0.472~ & 0.300~                                                          & 0.504~          & 0.328~                                                     & 0.555~          & 0.362~                                                      & 0.542~ & 0.316~                                                           & 0.620~ & 0.336~                                                               & 0.592~ & 0.338~                                                                  & 0.625~ & 0.383~                                                              & 0.621~ & 0.396~                                                                & 0.610~ & 0.376~                                                                & 0.624~ & 0.340~                                                              \\ 
\hline
\multirow{5}{*}{Exchange}    & 96  & 0.097~          & 0.209~                                                     & 0.083~ & \textbf{0.196~}                                                         & 0.100~ & 0.222~                                                                & 0.086~ & 0.206~                                                                   & 0.089~ & 0.210~                                                           & 0.083~ & 0.201~                                                          & \textbf{0.082~} & 0.199~                                                     & \textbf{0.082~} & 0.201~                                                      & 0.102~ & 0.235~                                                           & 0.107~ & 0.234~                                                               & 0.329~ & 0.440~                                                                  & 0.088~ & 0.218~                                                              & 0.085~ & 0.204~                                                                & 0.148~ & 0.237~                                                                & 0.111~ & 0.237~                                                              \\
                             & 192 & \textbf{0.103~} & \textbf{0.217~}                                            & 0.170~ & 0.291~                                                                  & 0.212~ & 0.326~                                                                & 0.177~ & 0.299~                                                                   & 0.182~ & 0.303~                                                           & 0.175~ & 0.296~                                                          & 0.176~          & 0.297~                                                     & 0.187~          & 0.307~                                                      & 0.172~ & 0.316~                                                           & 0.226~ & 0.344~                                                               & 0.544~ & 0.586~                                                                  & 0.176~ & 0.315~                                                              & 0.182~ & 0.303~                                                                & 0.271~ & 0.335~                                                                & 0.219~ & 0.335~                                                              \\
                             & 336 & \textbf{0.117~} & \textbf{0.232~}                                            & 0.316~ & 0.404~                                                                  & 0.379~ & 0.442~                                                                & 0.331~ & 0.417~                                                                   & 0.327~ & 0.414~                                                           & 0.327~ & 0.413~                                                          & 0.337~          & 0.418~                                                     & 0.345~          & 0.427~                                                      & 0.272~ & 0.407~                                                           & 0.367~ & 0.448~                                                               & 1.017~ & 0.786~                                                                  & 0.313~ & 0.427~                                                              & 0.348~ & 0.428~                                                                & 0.460~ & 0.476~                                                                & 0.421~ & 0.476~                                                              \\
                             & 720 & \textbf{0.158~} & \textbf{0.279~}                                            & 0.817~ & 0.679~                                                                  & 0.904~ & 0.715~                                                                & 0.847~ & 0.691~                                                                   & 0.852~ & 0.696~                                                           & 0.849~ & 0.694~                                                          & 0.909~          & 0.725~                                                     & 0.887~          & 0.708~                                                      & 0.714~ & 0.658~                                                           & 0.964~ & 0.746~                                                               & 1.239~ & 0.912~                                                                  & 0.839~ & 0.695~                                                              & 1.025~ & 0.774~                                                                & 1.195~ & 0.769~                                                                & 1.092~ & 0.769~                                                              \\ 
\cline{2-32}
                             & Avg & \textbf{0.119~} & \textbf{0.234 ~}                                           & 0.347~ & 0.393~                                                                  & 0.399~ & 0.426~                                                                & 0.360~ & 0.403~                                                                   & 0.363~ & 0.406~                                                           & 0.359~ & 0.401~                                                          & 0.376~          & 0.410~                                                     & 0.375~          & 0.411~                                                      & 0.315~ & 0.404~                                                           & 0.416~ & 0.443~                                                               & 0.782~ & 0.681~                                                                  & 0.354~ & 0.414~                                                              & 0.410~ & 0.427~                                                                & 0.519~ & 0.454~                                                                & 0.461~ & 0.454~                                                              \\ 
\hline
\multirow{5}{*}{ILI}         & 24  & \textbf{0.290~} & \textbf{0.311~}                                            & 1.701~ & 0.823~                                                                  & 2.122~ & 0.874~                                                                & 2.014~ & 0.899~                                                                   & 3.340~ & 1.299~                                                           & 2.585~ & 1.065~                                                          & 0.432~          & 0.487~                                                     & 1.724~          & 0.843~                                                      & 2.684~ & 1.112~                                                           & 2.317~ & 0.934~                                                               & 3.041~ & 1.186~                                                                  & 2.398~ & 1.040~                                                              & 2.527~ & 1.020~                                                                & 3.228~ & 0.945~                                                                & 2.294~ & 0.945~                                                              \\
                             & 36  & \textbf{0.305~} & \textbf{0.336~}~                                           & 1.941~ & 0.888~                                                                  & 2.289~ & 0.931~                                                                & 2.115~ & 0.943~                                                                   & 4.016~ & 1.453~                                                           & 2.623~ & 1.085~                                                          & 0.446~          & 0.479~                                                     & 1.536~          & 0.752~                                                      & 2.667~ & 1.068~                                                           & 1.972~ & 0.920~                                                               & 3.406~ & 1.232~                                                                  & 2.646~ & 1.088~                                                              & 2.615~ & 1.007~                                                                & 2.679~ & 0.848~                                                                & 1.825~ & 0.848~                                                              \\
                             & 48  & \textbf{0.311~} & \textbf{0.342~}                                            & 1.847~ & 0.893~                                                                  & 2.165~ & 0.908~                                                                & 2.188~ & 0.972~                                                                   & 4.541~ & 1.554~                                                           & 2.465~ & 1.035~                                                          & 0.477~          & 0.510~                                                     & 1.821~          & 0.832~                                                      & 2.558~ & 1.052~                                                           & 2.238~ & 0.940~                                                               & 3.459~ & 1.221~                                                                  & 2.614~ & 1.086~                                                              & 2.359~ & 0.972~                                                                & 2.622~ & 0.900~                                                                & 2.010~ & 0.900~                                                              \\
                             & 60  & \textbf{0.368~} & \textbf{0.390~}                                            & 1.976~ & 0.952~                                                                  & 2.085~ & 0.909~                                                                & 2.114~ & 0.968~                                                                   & 4.406~ & 1.517~                                                           & 2.386~ & 1.029~                                                          & 0.578~          & 0.564~                                                     & 1.923~          & 0.842~                                                      & 2.747~ & 1.110~                                                           & 2.027~ & 0.928~                                                               & 3.640~ & 1.305~                                                                  & 2.804~ & 1.146~                                                              & 2.487~ & 1.016~                                                                & 2.857~ & 0.963~                                                                & 2.178~ & 0.963~                                                              \\ 
\cline{2-32}
                             & Avg & \textbf{0.319~} & \textbf{0.345~}                                            & 1.866~ & 0.889~                                                                  & 2.165~ & 0.906~                                                                & 2.108~ & 0.946~                                                                   & 4.076~ & 1.456~                                                           & 2.515~ & 1.054~                                                          & 0.483~          & 0.510~                                                     & 1.751~          & 0.817~                                                      & 2.664~ & 1.086~                                                           & 2.139~ & 0.931~                                                               & 3.387~ & 1.236~                                                                  & 2.616~ & 1.090~                                                              & 2.497~ & 1.004~                                                                & 2.847~ & 0.914~                                                                & 2.077~ & 0.914~                                                              \\ 
\hline
\multicolumn{2}{c|}{Count}         & \multicolumn{2}{c|}{58}                                                      & \multicolumn{2}{c|}{5}                                                           & \multicolumn{2}{c|}{0}                                                         & \multicolumn{2}{c|}{0}                                                            & \multicolumn{2}{c|}{0}                                                    & \multicolumn{2}{c|}{0}                                                   & \multicolumn{2}{c|}{1}                                                       & \multicolumn{2}{c|}{1}                                                        & \multicolumn{2}{c|}{0}                                                    & \multicolumn{2}{c|}{0}                                                        & \multicolumn{2}{c|}{0}                                                           & \multicolumn{2}{c|}{0}                                                       & \multicolumn{2}{c|}{0}                                                         & \multicolumn{2}{c|}{0}                                                         & \multicolumn{2}{c}{0}                                                        \\
\hline
\end{tabular}}
\end{table*}

\textbf{Short-term forecasting.} 
Short-term forecasting results for the time series are presented in Table \ref{short}. Analysis of this table yields the following conclusions:
\begin{itemize}
\item The predictions generated by ACNet surpass those of other baseline models, with the Transformer-based Crossformer model achieving the second-best results. The success of ACNet can be attributed to its comprehensive feature extraction approach, effectively capturing and utilizing local, global, and nonlinear features. This significantly enhances the model's predictive performance. 
Crossformer also demonstrates its effectiveness in capturing complex patterns within real-world time series by explicitly extracting the interdependencies between different variables in the time series. 
\item Transformer-based models (e.g., iTransformer, PatchTST) exhibit limited capabilities in handling time series forecasting tasks involving nonlinear features. This limitation may stem from their reliance on the self-attention mechanism to capture global information from the input sequence. Nonlinear relationships, however, often require local contextual information for better understanding and accurate feature extraction.
\item The MICN model, which is based on multi-scale dilated convolution, shows the poorest performance across all datasets. This outcome suggests that traditional convolutions with fixed-size receptive fields are inadequate for real-world time series datasets containing numerous nonlinear features. The fixed nature of these receptive fields restricts their flexibility in handling nonlinear features of varying scales, leading to information redundancy and diminished predictive performance.
\end{itemize}

\begin{table*}[h]
\centering
\caption{Short-term multivariate forecasting results with different metrics. The best results are in bold numbers and the second best are highlighted with an underline. }
\label{short}
\resizebox{\columnwidth}{!}{
\begin{tabular}{c|c|c|c|c|c|c|c|c|c|c|c|c|c} 

\hline
\multicolumn{2}{c|}{Models}    & \begin{tabular}[c]{@{}c@{}}ACNet\\(Ours)\end{tabular} & \begin{tabular}[c]{@{}c@{}}ConvTimeNet\\(2024)\end{tabular} & \begin{tabular}[c]{@{}c@{}}TimeMixer\\(2024)\end{tabular} & \begin{tabular}[c]{@{}c@{}}iTransformer\\(2024)\end{tabular} & \begin{tabular}[c]{@{}c@{}}FITS\\(2024)\end{tabular} & \begin{tabular}[c]{@{}c@{}}PDF\\(2024)\end{tabular} & \begin{tabular}[c]{@{}c@{}}TSLANet\\(2024)\end{tabular} & \begin{tabular}[c]{@{}c@{}}TSMixer\\(2023)\end{tabular} & \begin{tabular}[c]{@{}c@{}}PatchTST\\(2023)\end{tabular} & \begin{tabular}[c]{@{}c@{}}MICN\\(2023)\end{tabular} & \begin{tabular}[c]{@{}c@{}}TimesNet\\(2023)\end{tabular} & \begin{tabular}[c]{@{}c@{}}Crossformer\\(2023)\end{tabular}  \\ 
\hline
\multirow{3}{*}{PEMS03} & MAE  & \textbf{0.145}                                          & 0.245                                                       & 0.187                                                     & 0.178                                                        & 0.348                                                & 0.195                                               & 0.185                                                   & 0.198                                                   & 0.212                                                    & 0.462                                                & 0.190                                                    & \uline{0.166}                                                \\
                        & MAPE & \textbf{1.295}                                          & 1.898                                                       & 1.466                                                     & 1.364                                                        & 2.656                                                & 2.794                                               & 1.459                                                   & 1.567                                                   & 1.636                                                    & 2.948                                                & 1.455                                                    & \uline{1.321}                                                \\
                        & RMSE & \textbf{0.208}                                          & 0.361                                                       & 0.277                                                     & 0.266                                                        & 0.468                                                & 0.293                                               & 0.278                                                   & 0.277                                                   & 0.310                                                    & 0.606                                                & 0.290                                                    & \uline{0.253}                                                \\ 
\hline
\multirow{3}{*}{PEMS04} & MAE  & \textbf{0.183}                                          & 0.307                                                       & 0.247                                                     & 0.235                                                        & 0.448                                                & 0.260                                               & 0.272                                                   & 0.238                                                   & 0.269                                                    & 0.935                                                & 0.269                                                    & \uline{0.206}                                                \\
                        & MAPE & \textbf{1.550}                                          & 2.635                                                       & 2.039                                                     & 1.931                                                        & 3.658                                                & 2.131                                               & 2.179                                                   & 1.985                                                   & 2.177                                                    & 7.799                                                & 2.208                                                    & \uline{1.682}                                                \\
                        & RMSE & \textbf{0.311}                                          & 0.566                                                       & 0.487                                                     & 0.481                                                        & 0.623                                                & 0.509                                               & 0.526                                                   & 0.434                                                   & 0.516                                                    & 1.298                                                & 0.534                                                    & \uline{0.430}                                                \\ 
\hline
\multirow{3}{*}{PEMS07} & MAE  & \textbf{0.130}                                          & 0.257                                                       & 0.187                                                     & 0.181                                                        & 0.382                                                & 0.194                                               & 0.197                                                   & 0.186                                                   & 0.211                                                    & 0.622                                                & 0.199                                                    & \uline{0.150}                                                \\
                        & MAPE & \textbf{1.349}                                          & 2.561                                                       & 1.907                                                     & 1.849                                                        & 3.509                                                & 2.004                                               & 1.948                                                   & 1.970                                                   & 2.058                                                    & 4.773                                                & 1.904                                                    & \uline{1.667}                                                \\
                        & RMSE & \textbf{0.191}                                          & 0.368                                                       & 0.279                                                     & 0.274                                                        & 0.501                                                & 0.290                                               & 0.297                                                   & 0.273                                                   & 0.303                                                    & 0.800                                                & 0.302                                                    & \uline{0.232}                                                \\ 
\hline
\multirow{3}{*}{PEMS08} & MAE  & \textbf{0.124}                                          & 0.313                                                       & 0.259                                                     & 0.255                                                        & 0.473                                                & 0.268                                               & 0.256                                                   & 0.259                                                   & 0.285                                                    & 0.797                                                & 0.302                                                    & \uline{0.238}                                                \\
                        & MAPE & 1.594                                                   & 2.115                                                       & 1.649                                                     & 1.705                                                        & 3.235                                                & 1.704                                               & 1.709                                                   & 1.685                                                   & 1.800                                                    & 5.239                                                & 2.077                                                    & \textbf{1.497}                                               \\
                        & RMSE & \textbf{0.204}                                          & 0.582                                                       & 0.523                                                     & 0.516                                                        & 0.723                                                & 0.531                                               & 0.517                                                   & 0.481                                                   & 0.540                                                    & 1.189                                                & 0.664                                                    & \uline{0.497}                                                \\
\hline
\end{tabular}}
\end{table*}

\subsection{Visualization of forecasting} 
\begin{figure}[h]
	\centering
	\includegraphics[width=0.7\columnwidth]{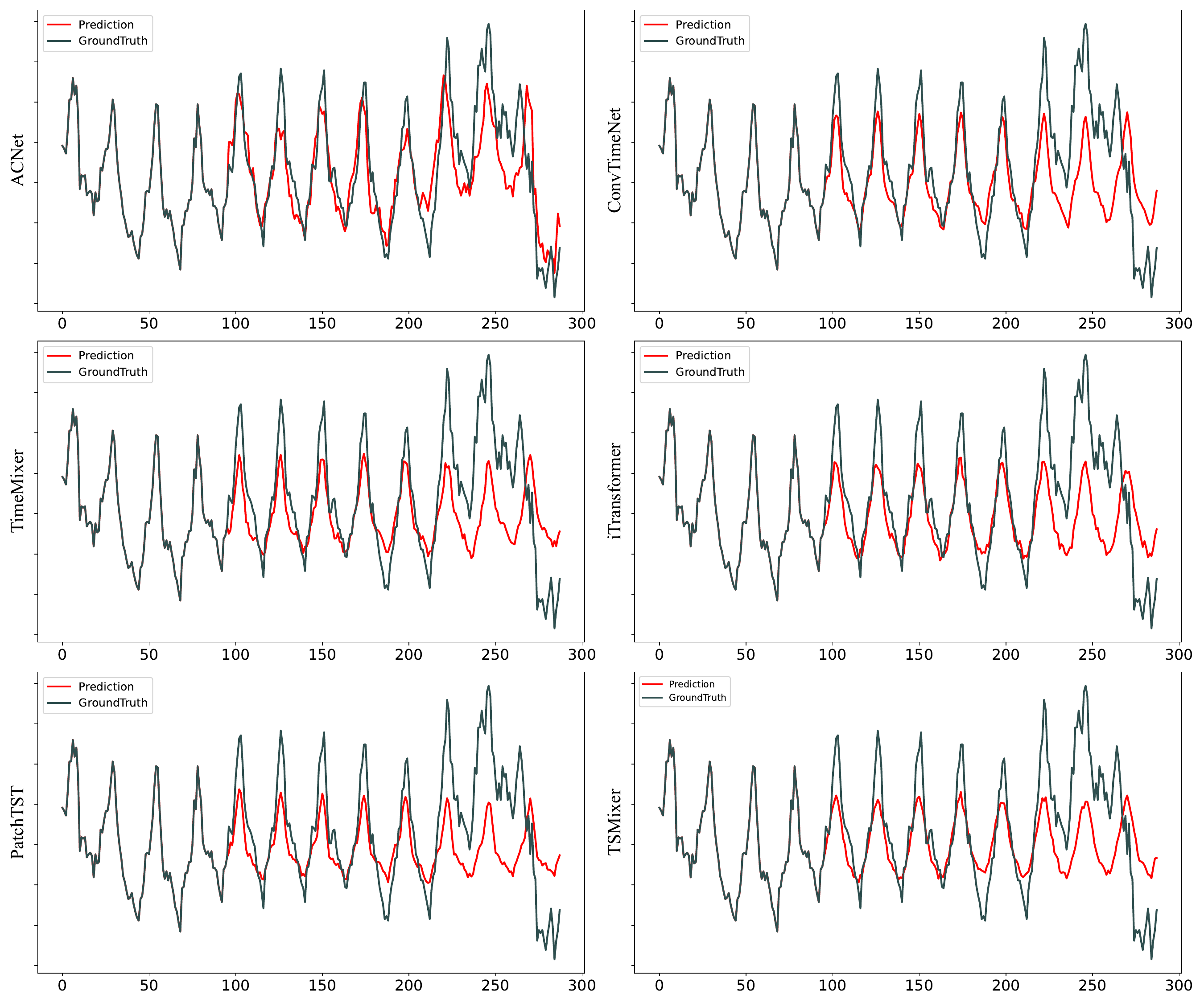} 
	\caption{Visualize the ETTh1 dataset, where the first 96 data points represent the input length.}\label{show}
\end{figure}

To evaluate the predictive performance of ACNet compared to SOTA TSF models (such as ConvTimeNet, TimeMixer, iTransformer, TSMixer and PatchTST) in real-world scenarios, we qualitatively compared the prediction results of the last dimension on the test set of the ETTh1 dataset, as shown in Figure \ref{show}. The input length was set to 96, and the prediction length was set to 192 to assess the fitting between the predicted sequences and the actual sequences.

From Figure \ref{show}, it can be observed that ACNet is capable of adapting to the dynamic changes in time series data, capturing both the periodic and trend information. However, while other models effectively capture the periodic information of the time series, they neglect to acquire trend information. For instance, around the 280th time point, the sequence suddenly oscillates downward. ACNet successfully captured the trend information of such non-stationary oscillatory changes, while other models retained the trend information from the previous time period, indicating that their predictive performance did not benefit from the module that extracts long-term feature information. In summary, ACNet can not only capture local contextual information but also effectively capture the global trend information and nonlinear information of time series, thereby achieving better fitting results.

\subsection{Long-term information utilization} \label{long}

\begin{figure}[h]
	\centering
	\includegraphics[width=0.7\columnwidth]{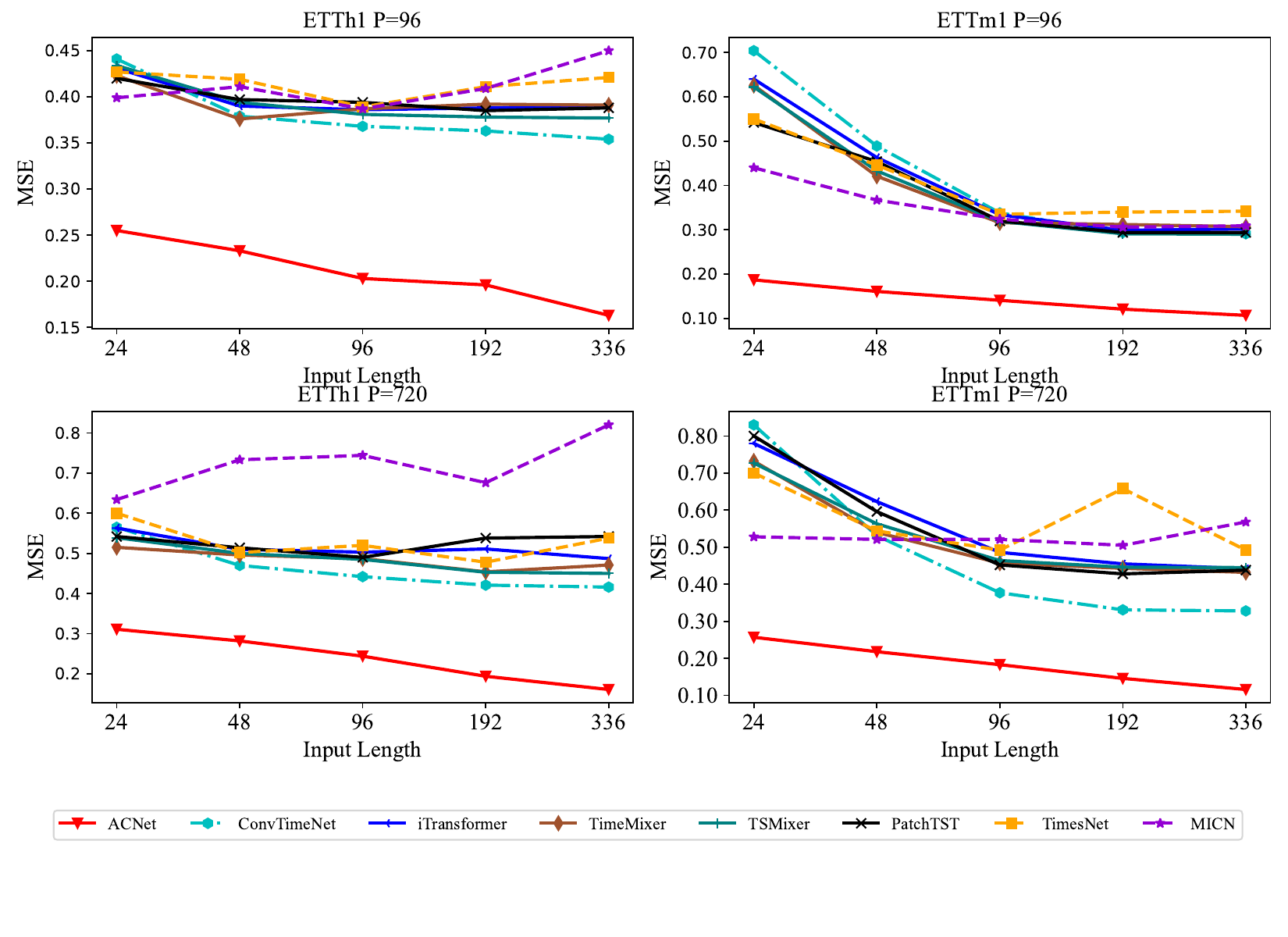} 
	\caption{The predictive performance (MSE) of different input sequence windows.}\label{look_back}
\end{figure}

To validate ACNet's ability to capture long-term correlations, we compared the performance of different models on the ETTh1 and ETTm1 datasets, with output lengths set to $\left \{ 96,720 \right \}$ and input sequence windows set to $\left \{ 24, 48, 96, 192, 336 \right \}$. In theory, using longer input sequence windows can increase the receptive field and potentially improve prediction performance. However, as evidenced by the experimental results in Figure \ref{look_back}, the performance of Transformer-based models, TimeMixer, and MICN models does not continuously improve with increasing model input length, indicating that these models do not benefit from longer input sequence windows, i.e., they lack in capturing long-term correlations. In contrast, ACNet, ConvTimeNet, and TSMixer show continuous improvement in model performance (decreasing MSE values) with increasing input sequence length, demonstrating that these models can exploit more features from longer input sequence windows and effectively capture the long-term correlations in the input sequence.

\subsection{T-tests}\label{tests}

\begin{figure}[h]
	\centering
	\includegraphics[width=0.65\columnwidth]{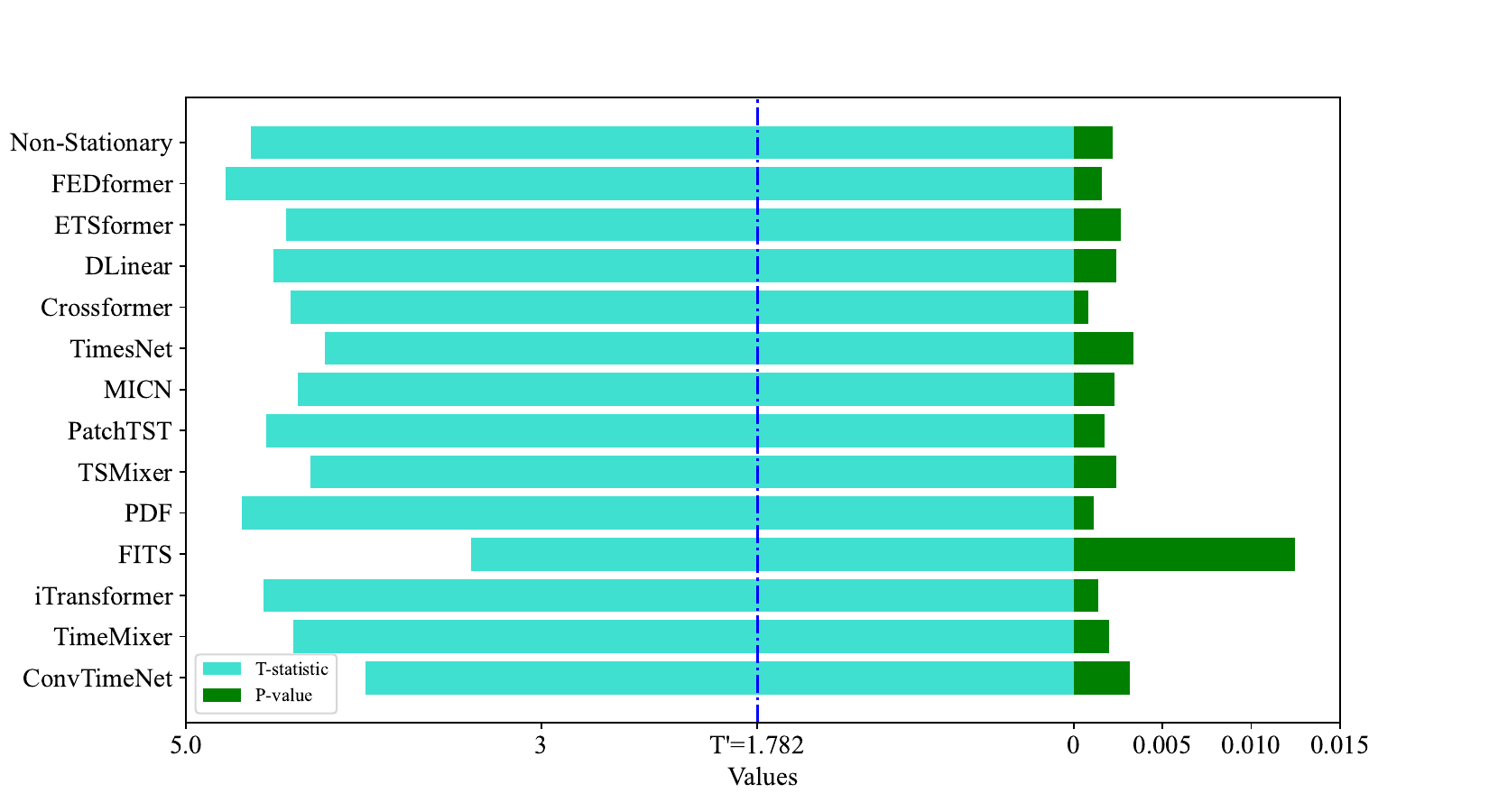} 
	\caption{The result of the statistical significance test between ACNet and fourteen baseline models.}\label{t-test}
\end{figure}

To further validate the effectiveness of the ACNet model compared to other baseline models, we conducted T-tests to assess the significant differences between the prediction results of ACNet and the baseline models. As shown in Figure \ref{t-test}, we can draw the following conclusions: (1) The T-statistic for ACNet relative to all baseline models significantly exceeds $T' = 1.782$ (by looking up the critical value in the T-distribution table based on the P-value, we obtain the T-value), indicating that the average MSE of ACNet is significantly lower than that of the baseline models. (2) The calculated P-values are all less than the threshold value of $P = 0.05$, indicating statistically significant differences in prediction results among different models. Based on the above analysis, we can conclude that the observed performance differences between ACNet and the baseline models are indeed the result of true differences, rather than random factors.

\subsection{Validation of capturing nonlinear features}\label{nonlinear}
To further verify the effectiveness of our proposed nonlinear feature adaptive extraction module in extracting features from real-world time series data, we selected the Exchange dataset for validation. We processed the data using both standard convolutional kernels and our proposed gated deformable convolution (GDC). We used a batch of the dataset and extracted feature maps from the intermediate layers to observe the activation of input data after passing through the convolutional layers. This helps in understanding how different convolutional modules capture nonlinear relationships. We visualized the feature maps from the first eight layers, as shown in Figure \ref{nonlinearF}:

\begin{figure}[!htbp]
	\centering
	\subfigure[Standard Converlutional Feature Map.]{
		\includegraphics[width=0.98\columnwidth, height=4.6cm]{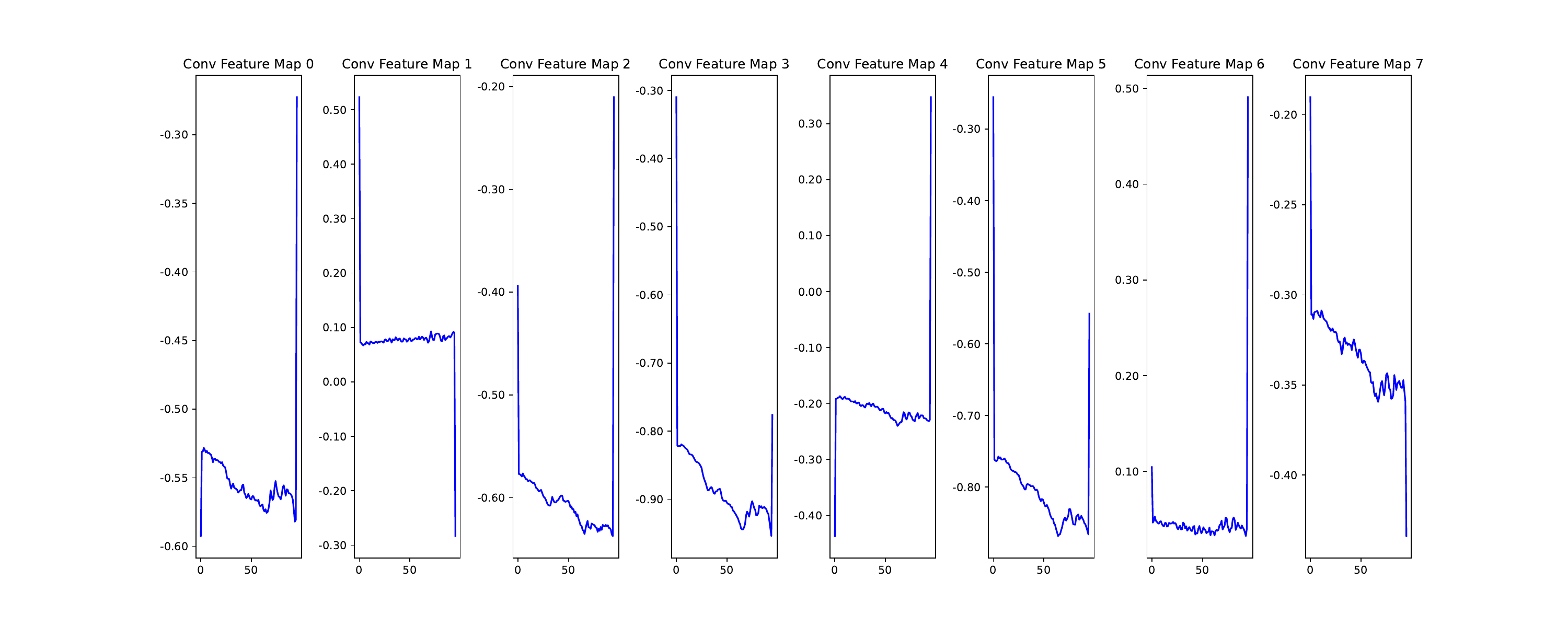}
	}
	\subfigure[GDC Feature Map.]{
		\includegraphics[width=0.98\columnwidth, height=4.6cm]{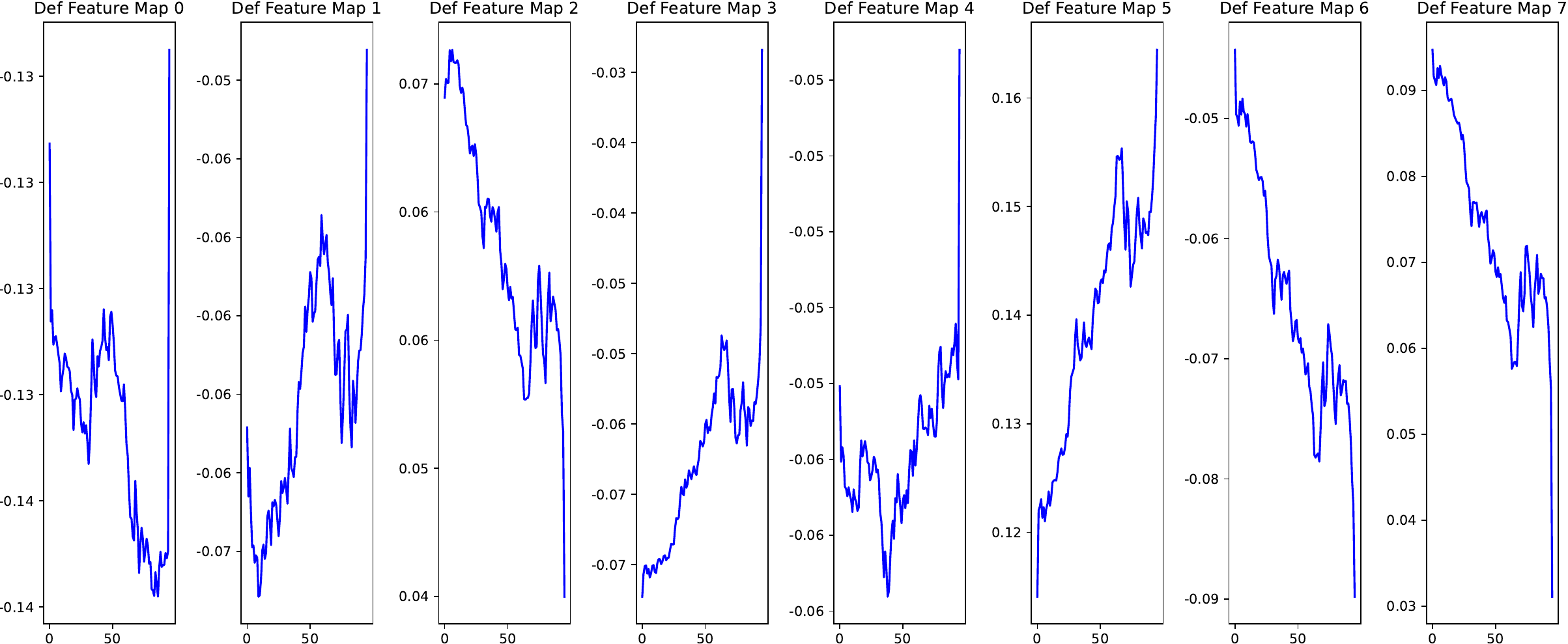}
	}
	\caption{Visualization of feature maps from different convolutions.}
	\label{nonlinearF}
\end{figure}

From Figure \ref{nonlinearF}, we can draw the following conclusions:

\textbf{(1) Amplitude of Variations in Feature Maps.}
The feature maps from standard convolution exhibit smaller amplitudes of change with fewer fluctuations, indicating a potential inadequacy in handling sudden and nonlinear changes in the input data.
The feature maps from GDC exhibit larger amplitudes of change with more noticeable fluctuations, indicating a greater flexibility in capturing sudden and nonlinear changes in the input data.

\textbf{(2) Diversity in Feature Maps.}
The feature maps from standard convolution show more consistent patterns with relatively smooth changes.
The feature maps from GDC show more diverse patterns and variations, indicating a stronger adaptability in handling nonlinear features.

\textbf{(3) Response to Sudden Changes.}
The feature maps from standard convolution exhibit smooth transitions at points of sudden changes in the data, lacking sensitivity to such changes.
The feature maps from GDC exhibit noticeable fluctuations and variations at points of sudden changes in the data, better capturing these sudden features.

Specifically, from the standard convolutional feature maps (as shown in Figure \ref{nonlinearF}(a)), we can see that most feature maps show relatively smooth and stable changes with almost no significant fluctuations or sudden changes. In feature map 5, some fluctuations can be observed, but the overall changes remain relatively stable.
From the GDC feature maps (as shown in Figure \ref{nonlinearF}(b)), we can see that feature maps 1, 3, 4, and 5 exhibit significant fluctuations and sudden changes, indicating that GDC can capture dynamic sudden features in the input data. Feature maps 0 and 2 exhibit more details and fluctuations, suggesting that GDC performs well in handling non-stationary features.

In summary, GDC performs better in handling nonlinear features because it can dynamically adjust the positions of the convolutional kernels, making it more adaptable to variations in the input data.

\subsection{Efficiency analysis}\label{overhead}
To evaluate the efficiency of ACNet, we selected the ETTh1 dataset and compared it with six SOTA models. We conducted efficiency analysis of the models from four aspects: computational complexity (Flops), model size (Parameters), training time, and inference time. To facilitate a clearer comparison and analysis of model efficiency, we integrated these four metrics and computed their average values as the model efficiency scores, as shown in Table \ref{efficitncy}. Table \ref{efficitncy} distinctly demonstrates the efficiency advantages of ACNet, which are particularly crucial for practical applications of TSF.
This advantage mainly stems from the model's parameter sharing mechanism, deformable convolutions, and the ability to rapidly update network weights through pseudo-inverse algorithms.
Additionally, we observed a decreasing trend in the training time of iTransformer with increasing input length. This is because iTransformer employs an early termination strategy during training, where training halts when the loss value decreases continuously for a certain number of iterations (n=3 in iTransformer). According to our experiments, as the input length increases, the model is more likely to meet this condition during training.

\begin{table}
\centering
\caption{Results of the efficiency analysis.}\label{efficitncy}
\resizebox{\columnwidth}{!}{
\begin{tabular}{c|c|c|c|c|c|c|c} 
\hline
\multicolumn{2}{c|}{Metrics}         & \multirow{2}{*}{\begin{tabular}[c]{@{}c@{}}Flops\\(G)\end{tabular}} & \multirow{2}{*}{\begin{tabular}[c]{@{}c@{}}Parameters\\(M)\end{tabular}} & \multirow{2}{*}{\begin{tabular}[c]{@{}c@{}}Training Time \\(s)\end{tabular}} & \multirow{2}{*}{\begin{tabular}[c]{@{}c@{}}Inference Time \\(s)\end{tabular}} & \multicolumn{2}{c}{Ranking}                           \\ 
\cline{1-2}\cline{7-8}
\multicolumn{2}{c|}{Models}          &                                                                     &                                                                          &                                                                              &                                                                               & Four Metrics                & Avg Ranking             \\ 
\hline
\multirow{4}{*}{ACNet}       & 192  & 0.064~                                                              & 0.949~                                                                   & 20.308~                                                                      & 0.091                                                                         & \multirow{4}{*}{(1,4,3,1)}  & \multirow{4}{*}{2.25~}  \\
                              & 384  & 0.128~                                                              & 1.898~                                                                   & 35.999~                                                                      & 0.124                                                                         &                             &                         \\
                              & 768  & 0.254~                                                              & 3.700~                                                                   & 49.618~                                                                      & 0.217                                                                         &                             &                         \\
                              & 1536 & 0.510~                                                              & 7.589~                                                                   & 73.758~                                                                      & 0.436                                                                         &                             &                         \\ 
\hline
\multirow{4}{*}{ConvTimeNet}  & 192  & 4.745~                                                              & 0.265~                                                                   & 14.978~                                                                      & 7.977                                                                         & \multirow{4}{*}{(4,1,2,7)} & \multirow{4}{*}{3.50~}  \\
                              & 384  & 9.491~                                                              & 0.335~                                                                   & 24.781~                                                                      & 8.975                                                                         &                             &                         \\
                              & 768  & 18.981~                                                             & 0.476~                                                                   & 38.070~                                                                      & 10.97                                                                         &                             &                         \\
                              & 1536 & 37.962~                                                             & 0.757~                                                                   & 48.436~                                                                      & 13.962                                                                        &                             &                         \\ 
\hline
\multirow{4}{*}{iTransformer} & 192  & 0.305~                                                              & 0.866~                                                                   & 1056.093~                                                                    & 2.493                                                                         & \multirow{4}{*}{(2,2,7,3)} & \multirow{4}{*}{3.50~}  \\
                              & 384  & 0.322~                                                              & 0.915~                                                                   & 1036.082~                                                                    & 2.658                                                                         &                             &                         \\
                              & 768  & 0.356~                                                              & 1.014~                                                                   & 422.610~                                                                     & 2.986                                                                         &                             &                         \\
                              & 1536 & 0.426~                                                              & 1.210~                                                                   & 407.591~                                                                     & 3.155                                                                         &                             &                         \\ 
\hline
\multirow{4}{*}{TimeMixer}    & 192  & 4.329~                                                              & 0.245~                                                                   & 14.209~                                                                      & 6.979                                                                         & \multirow{4}{*}{(6,5,1,6)}  & \multirow{4}{*}{4.50~}  \\
                              & 384  & 14.895~                                                             & 0.902~                                                                   & 17.310~                                                                      & 7.978                                                                         &                             &                         \\
                              & 768  & 54.757~                                                             & 3.464~                                                                   & 27.918~                                                                      & 8.576                                                                         &                             &                         \\
                              & 1536 & 209.391~                                                            & 13.571~                                                                  & 54.004~                                                                      & 9.474                                                                         &                             &                         \\ 
\hline
\multirow{4}{*}{TSMixer}      & 192  & 0.027                                                               & 0.100                                                                    & 38.392~                                                                      & 1.723                                                                         & \multirow{4}{*}{(3,3,4,2)}  & \multirow{4}{*}{3.00~}  \\
                              & 384  & 0.087                                                               & 0.339                                                                    & 41.811~                                                                      & 1.931                                                                         &                             &                         \\
                              & 768  & 0.306                                                               & 1.239                                                                    & 56.190~                                                                      & 1.976                                                                         &                             &                         \\
                              & 1536 & 1.139                                                               & 4.727                                                                    & 69.689~                                                                      & 2.200                                                                         &                             &                         \\ 
\hline
\multirow{4}{*}{PatchTST}     & 192  & 17.253~                                                             & 4.140~                                                                   & 51.121~                                                                      & 3.649                                                                         & \multirow{4}{*}{(5,6,5,4)}  & \multirow{4}{*}{5.00~}  \\
                              & 384  & 34.505~                                                             & 5.265~                                                                   & 80.290~                                                                      & 3.936                                                                         &                             &                         \\
                              & 768  & 69.010~                                                             & 7.515~                                                                   & 87.804~                                                                      & 4.057                                                                         &                             &                         \\
                              & 1536 & 138.021~                                                            & 12.015~                                                                  & 101.786~                                                                     & 3.955                                                                         &                             &                         \\ 
\hline
\multirow{4}{*}{MICN}         & 192  & 108.741                                                             & 27.542                                                                   & 116.441~                                                                     & 2.926                                                                         & \multirow{4}{*}{(7,7,6,5)}  & \multirow{4}{*}{6.25~}  \\
                              & 384  & 187.220                                                             & 36.559                                                                   & 143.929~                                                                     & 3.174                                                                         &                             &                         \\
                              & 768  & 364.311                                                             & 48.594                                                                   & 174.319~                                                                     & 3.823                                                                         &                             &                         \\
                              & 1536 & 799.024                                                             & 76.665                                                                   & 273.936~                                                                     & 5.757                                                                         &                             &                         \\
\hline
\end{tabular}}
\end{table}

\subsection{Ablation study}\label{ab}


To assess the effectiveness of each component within ACNet, we conducted the following ablations: (1) w/o GDC: remove the gated deformable convolution module used for extracting nonlinear features of time series; (2) w/o Temporal: elimination of modules for extracting temporal correlation information; (3) w/o ALL: complete removal of all feature extraction modules. We conducted ablation studies on the ETTh2 and Electricity datasets, with input lengths set to 96 and prediction lengths set to $\left\{ 96, 192, 336, 720\right\}$. 

The results in Table \ref{ablation} indicate that removing any module from the original model led to a decrease in performance. Specifically: (1) Removing the GDC module (w/o GDC) will cause the model to lose its ability to extract nonlinear information from the time series. By solely relying on the the time-domain information extracted by the model, the prediction accuracy of the model is reduced. (2) Removing the temporal feature extraction module (w/o temporal) will prevent the model from extracting local contextual information and global patterns from the time series, leading to a decline in predictive performance. (3) Completely removing the feature extraction module (w/o ALL) is equivalent to using only the FFN layer for prediction. Experimental results indicate that this approach performs poorly, once again demonstrating that FFN lacks the ability to effectively extract complex features from time series data.


\begin{table}
\centering
\caption{Results of the ablation study.}
\label{ablation}
\begin{tabular}{c|c|lccccccccccc} 
\hline
\multicolumn{2}{c|}{Models}        & \multicolumn{1}{c}{} & \multicolumn{2}{c}{ACNet}       &  & \multicolumn{2}{c}{w/o GDC} &  & \multicolumn{2}{c}{w/o Temporal} &  & \multicolumn{2}{c}{w/o All}  \\ 
\hline
\multicolumn{2}{c|}{Metric}        & \multicolumn{1}{c}{} & MSE             & MAE             &  & MSE    & MAE                     &  & MSE    & MAE                     &  & MSE    & MAE                 \\ 
\hline
\multirow{5}{*}{ETTh2}       & 96  &                      & 0.147~          & 0.259~          &  & 0.156~ & 0.267~                  &  & 0.171~ & 0.276~                  &  & 0.185~ & 0.287~              \\
                             & 192 &                      & 0.168~          & 0.279~          &  & 0.180~ & 0.287~                  &  & 0.203~ & 0.300~                  &  & 0.226~ & 0.315~              \\
                             & 336 &                      & 0.192~          & 0.295~          &  & 0.205~ & 0.304~                  &  & 0.236~ & 0.319~                  &  & 0.269~ & 0.337~              \\
                             & 720 &                      & 0.223~          & 0.310~          &  & 0.237~ & 0.320~                  &  & 0.277~ & 0.340~                  &  & 0.325~ & 0.363~              \\ 
\cline{2-14}
                             & Avg &                      & \textbf{0.183~} & \textbf{0.286~} &  & 0.195~ & 0.295~                  &  & 0.222~ & 0.309~                  &  & 0.251~ & 0.326~              \\ 
\hline
\multirow{5}{*}{Electricity} & 96  &                      & 0.131~          & 0.233~          &  & 0.136~ & 0.238~                  &  & 0.141~ & 0.243~                  &  & 0.144~ & 0.246~              \\
                             & 192 &                      & 0.135~          & 0.239~          &  & 0.144~ & 0.249~                  &  & 0.143~ & 0.247~                  &  & 0.146~ & 0.250~              \\
                             & 336 &                      & 0.138~          & 0.245~          &  & 0.153~ & 0.255~                  &  & 0.148~ & 0.253~                  &  & 0.151~ & 0.257~              \\
                             & 720 &                      & 0.142~          & 0.250~          &  & 0.155~ & 0.257~                  &  & 0.150~ & 0.255~                  &  & 0.153~ & 0.258~              \\ 
\cline{2-14}
                             & Avg &                      & \textbf{0.137~} & \textbf{0.242~} &  & 0.147~ & 0.250~                  &  & 0.146~ & 0.250~                  &  & 0.149~ & 0.253~              \\
\hline
\end{tabular}
\end{table}

\section{Conclusion}\label{sec:sample5}
This paper introduces a feature-driven time series prediction network, ACNet. From the perspective of effective feature extraction and utilization, we use convolutional kernels with different dilation factors to capture latent local patterns and features at different resolutions in complex time series. Additionally, adaptive average pooling is employed to learn the global temporal patterns of the time series. By utilizing improved gated mechanism-based deformable convolutions, the network can effectively capture nonlinear features and pattern information in the time series.
Considering the inherent variability in the data distribution of time series, the model allows for dynamic updates based on changes in the data. Through extensive experimentation on twelve benchmark datasets, our findings demonstrate the superior performance of ACNet compared to previous SOTA models.

\section*{Acknowledgments}

This work is partially supported by National Hi-Tech Project, China with grant No. WDZC20215250117.



\printcredits

\bibliographystyle{model1-num-names}
\bibliography{cas-refs}


\end{document}